\documentclass[5p]{elsarticle}
\usepackage{graphicx}
\usepackage{amsmath,amsfonts,amssymb}
\usepackage{bm}
\usepackage{algorithm}
\usepackage{algorithmic}
\usepackage{afterpage}
\usepackage{url}
\usepackage[breaklinks=true,letterpaper=true,colorlinks,bookmarks=false]{hyperref}
\usepackage[table]{xcolor}
\usepackage{multirow}

\usepackage[pagewise]{lineno}

\biboptions{sort}

\definecolor{deep_blue}{rgb}{0,.2,.5}
\definecolor{dark_blue}{rgb}{0,.15,.5}

\hypersetup{pdftex,  
  breaklinks=true,  
  colorlinks=true,
  linkcolor=dark_blue,
  citecolor=deep_blue,
}

\def\B#1{\mathbf{#1}}

\def\slantfrac#1#2{\kern.1em^{#1}\kern-.1em/\kern-.1em_{#2}}
\newlength{\mylength}%

\newif\iffinal
\finaltrue
\iffinal\else
\usepackage[tightpage,active]{preview}
\fi

\usepackage{xcolor}
\usepackage[normalem]{ulem}
\colorlet{markercolor}{purple!50!black}

\usepackage{MnSymbol}
\begin{document}



\title{Assessing and tuning brain decoders: cross-validation, caveats, and
guidelines}

\author[parietal,cea]{Ga\"el Varoquaux\corref{corresponding}}
\author[uoft,rotman]{Pradeep Reddy Raamana}
\author[unicog,cea,icm]{Denis A. Engemann}
\author[parietal,cea]{Andr\'es Hoyos-Idrobo}
\author[parietal,cea]{Yannick Schwartz}
\author[parietal,cea]{Bertrand Thirion}

\cortext[corresponding]{Corresponding author}

\address[parietal]{Parietal project-team, INRIA Saclay-\^ile de France,
France}
\address[cea]{CEA/Neurospin b\^at 145, 91191 Gif-Sur-Yvette, France}
\address[rotman]{Rotman Research Institute, Baycrest Health Sciences, Toronto, ON, Canada M6A 2E1}
\address[uoft]{Dept. of Medical Biophysics, University of Toronto, Toronto, ON, Canada M5S 1A1}
\address[unicog]{Cognitive Neuroimaging Unit, INSERM, Universit\'e Paris-Sud and
Universit\'e Paris-Saclay, 91191 Gif-sur-Yvette, France}
\address[icm]{Neuropsychology \& Neuroimaging team
INSERM UMRS 975, Brain and Spine Institute (ICM), Paris}

\begin{abstract}
Decoding, \emph{ie} prediction from brain images or signals, calls for
empirical evaluation of its predictive power. Such evaluation is achieved via
cross-validation, a method also used to tune decoders' hyper-parameters.
This paper is a review on cross-validation procedures
for decoding in neuroimaging.
It includes a didactic overview of the relevant theoretical
considerations. Practical aspects are highlighted with an extensive
empirical study of the common decoders in within- and across-subject
predictions, on multiple datasets --anatomical and functional MRI and MEG-- and simulations.
Theory and experiments outline that the popular ``leave-one-out''
strategy leads to unstable and biased estimates, and a repeated random splits method should be
preferred.
Experiments outline the large error bars of
cross-validation in neuroimaging settings: typical confidence intervals of 10\%. Nested cross-validation can tune decoders' parameters while
avoiding circularity bias. However we find that it can be more favorable to
use sane defaults, in particular for non-sparse decoders.
\end{abstract}

\begin{keyword}
    cross-validation; decoding; fMRI; model selection;
    sparse; bagging; MVPA
\end{keyword}

\maketitle

\sloppy 

\section{Introduction: decoding needs model evaluation}


Decoding, \emph{ie} predicting behavior or phenotypes from brain images
or signals,
has become a central tool in neuroimage data processing
\cite{haynes2006,haynes2015primer,kamitani2005decoding,norman2006,varoquaux2014machine,yarkoni2016choosing}.
In clinical applications, prediction opens the door to diagnosis or
prognosis \cite{mouraomiranda2005,fu2008pattern,demirci2008review}. To
study cognition, successful prediction is seen as evidence of a
link between observed behavior and a brain region \cite{haxby2001} or a
small fraction of the image \cite{kriegeskorte2006}. Decoding power can
test if an encoding model describes well 
multiple facets of stimuli \cite{mitchell2008,naselaris2011}.
Prediction can be used to establish what specific brain functions
are implied by observed activations \cite{schwartz2013,poldrack2009decoding}. 
All these applications rely on measuring the predictive power of a decoder.

%

Assessing predictive power is difficult as it calls for
characterizing the decoder on prospective data, rather than on the data
at hand. Another challenge is that the decoder must often choose between
many different estimates that give rise to the same prediction error on
the data, when there are more features (voxels) than samples
(brain images, trials, or subjects). For this choice, it relies on some form of regularization, that
embodies a prior on the solution \cite{hastie2009elements}.
The amount of regularization
is a parameter of the decoder that may require tuning. Choosing a
decoder, or setting appropriately its internal parameters, are important
questions for brain mapping, as these choice will not only condition the
prediction performance of the decoder, but also the brain features that it
highlights.

\smallskip


Measuring prediction accuracy is central to decoding, to assess a decoder,
select one in various alternatives, or tune its parameters.
The topic of this paper is cross-validation, the standard
tool to measure predictive power and tune parameters in decoding.
The first section is a primer on cross-validation giving the theoretical
underpinnings 
and the current practice in neuroimaging.
In the second section, we perform an extensive empirical study. This
study shows
that cross-validation results carry a large uncertainty, that
cross-validation should be performed on full blocks of correlated data, and
that repeated random splits should be preferred to leave-one-out.
Results also yield guidelines for decoder parameter choice
in terms of prediction performance and
stability.


\section{A primer on cross-validation}


This section is a tutorial introduction to important concepts in
cross validation for decoding from brain images.

\subsection{Cross-validation: estimating predictive power}

In neuroimaging, a decoder is a predictive model that, given brain images $\B{X}$, infers
an external variable $\B{y}$. Typically,
$\B{y}$ is a categorical variable giving the experimental condition or
the health status of subjects. The accuracy, or predictive
power, of this model is the expected error on the prediction, formally:
\begin{equation}
    \text{accuracy} = \mathbb{E}\bigl[
	\mathcal{E}(\B{y}^\text{pred}, \B{y}^\text{ground truth})
    \bigr]
    \label{eq:accuracy}
\end{equation}
where $\mathcal{E}$ is a measure of the error, most often\footnote{For
multi-class problems, where there is more than 2 categories in $\B{y}$,
or for unbalanced classes, a more elaborate choice is advisable, to
distinguish misses and false detections for each class.}
the fraction of instances for which $\B{y}^\text{pred} \neq
\B{y}^\text{ground truth}$. Importantly, in equation (\ref{eq:accuracy}),
$\mathbb{E}$ denotes the \emph{expectation}, \emph{ie} the average error
that the model would make on infinite amount of data generated from
the same experimental process. 

In decoding settings, the
investigator has access to labeled data, \emph{ie} brain images for which
the variable to predict, $\B{y}$, is known.
These data are used to train the model,
fitting the model parameters, and to estimate its predictive power. However,
the same observations cannot be used for both. Indeed, it is much easier
to find the correct labels for brain images that have been seen by the
decoder than for unknown images\footnote{A simple strategy that makes no
errors on
seen images is simply to store all these images during the training and,
when asked to predict on an image, to look up the corresponding label
in the store.}. The challenge is to measure the ability to
\emph{generalize} to new data.

\begin{figure}[t]
\hfill%
\begin{minipage}[T]{.45\linewidth}%
    \caption{\textbf{Cross-validation}: the data is split
    multiple times into a train
    set, used to train the model, and a test set, used to compute
    predictive power.
    \label{fig:cross_val}}%
\end{minipage}
\hfill
\begin{minipage}[T]{.45\linewidth}%
    \includegraphics[width=\linewidth]{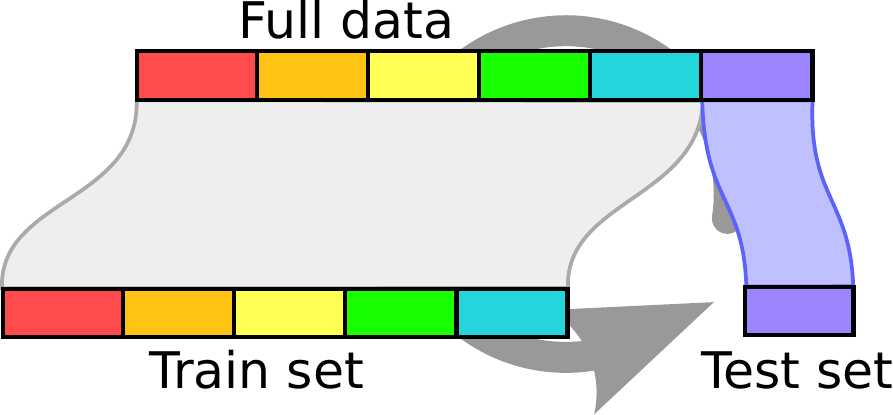}
\end{minipage}
\end{figure}

The standard approach to measure predictive power is
\emph{cross-validation}: the available data is split into a \emph{train
set}, used to train the model, and a \emph{test set}, unseen by the model
during training and used to compute a prediction error (figure
\ref{fig:cross_val}). Chapter 7 of \cite{hastie2009elements} contains a reference
on statistical aspects of cross-validation. Below, we detail important
considerations in neuroimaging.

\paragraph{Independence of train and test sets}

Cross-validation relies on independence between the train and test
sets. With time-series, as in fMRI, the autocorrelation of brain signals
and the temporal structure of the confounds imply that a time separation
is needed to give truly independent observations. In addition, to give
a meaningful estimate of prediction power, the test set should contain new
samples displaying all confounding uncontrolled sources of variability. 
For
instance, in multi-session data, it is harder to predict on a new session
than to leave out part of each session and use these samples as a test
set. However, generalization to new sessions is useful to capture actual invariant information.
Similarly, for multi-subject data, predictions on new subjects give results
that hold at the population level. However, a confound such as movement
may correlate with the diagnostic status predicted. In such a case the
amount of movement should be balanced between train and test set.

\paragraph{Sufficient test data}

Large test sets are necessary to obtain sufficient power for the prediction
error for each split of cross-validation. As the amount of data is limited, there
is a balance to strike between achieving such large test sets and keeping enough training
data to reach a good fit with the decoder. Indeed, theoretical results
show that cross-validation has a negative bias 
on small dataset \cite[sec.5.1]{arlot2010} as it involves fitting models on
a fraction of the data. On
the other hand, large test sets decrease
the variance of the estimated accuracy 
\cite[sec.5.2]{arlot2010}. A good cross-validation strategy
balances these two opposite effects. Neuroimaging papers often
use \emph{leave one out} cross-validation, leaving out a single
sample at each split. While this provides ample data for training, it
maximizes test-set variance
and does not yield stable estimates of predictive accuracy\footnote{One simple
aspect of the shortcomings of small test sets
is that they produce unbalanced dataset,
in particular
leave-one-out for which there is only one class
represented in the test set.}.
From a decision-theory
standpoint, it is preferable to leave out 10\% to 20\% of the data, as in
10-fold cross-validation \cite[chap.\,7.12]{hastie2009elements}
\cite{breiman1992submodel,kohavi1995study}. Finally, it is also beneficial to increase
the number of splits while keeping a given ratio between train 
and test set size. For this purpose k-fold can be replaced by strategies
relying on repeated random splits of the data
(sometimes called repeated learning-testing\footnote{Also related is
bootstrap CV, which may however duplicate samples inside the training set
of the test set.}
\cite{arlot2010} or \emph{ShuffleSplit}
\cite{pedregosa2011}). As discussed above, such splits should be consistent with the
dependence structure across the observations (using \emph{eg} a
\emph{LabelShuffleSplit}), or the training set could be stratified to avoid class imbalance \cite{Raamana2014ThickNetFusion_NBA}. In neuroimaging, good
strategies often involve leaving out sessions or 
subjects.

%

\subsection{Hyper-parameter selection}

\paragraph{A necessary evil: one size does not fit all}
In standard statistics, fitting a simple model on abundant data can be
done without the tricky choice of a meta-parameter: all model parameters
are estimated from the data, for instance with a maximum-likelihood
criterion. However, in high-dimensional settings, when the number of
model parameters is much larger than the sample size, some
form of regularization is needed. Indeed, adjusting model parameters to
best fit the data without restriction leads to \emph{overfit}, \emph{ie}
fitting noise \cite[chap.\,7]{hastie2009elements}. Some form of
regularization or prior is then necessary to 
restrict model complexity, \emph{e.g.} with low-dimensional PCA
in discriminant analysis \cite{chen2006exploring}, or by selecting a small
number of voxels with a sparse penalty \cite{yamashita2008,carroll2009}.
If too much
regularization is imposed, the ensuing models are too constrained by the prior, they
\emph{underfit}, \emph{ie} they do not exploit the full richness of the
data. Both underfitting and overfitting are detrimental to predictive
power and to the estimation of model weights, the decoder maps. 
Choosing the amount of regularization is a typical bias-variance problem: erring
on the side of variance leads to overfit, while too much bias leads to
underfit. In general, the best tradeoff is a data-specific choice,
governed by the statistical power of the prediction task: 
the amount of data and the
signal-to-noise ratio.


\begin{figure}[t]
\hfill%
\begin{minipage}[T]{.3\linewidth}%
    \caption{\textbf{Nested cross-validation}: two cross-validation loops
    are run one inside the other.
    \label{fig:nested_cross_val}}%
\end{minipage}
\hfill%
\begin{minipage}[T]{.68\linewidth}%
    \includegraphics[width=\linewidth]{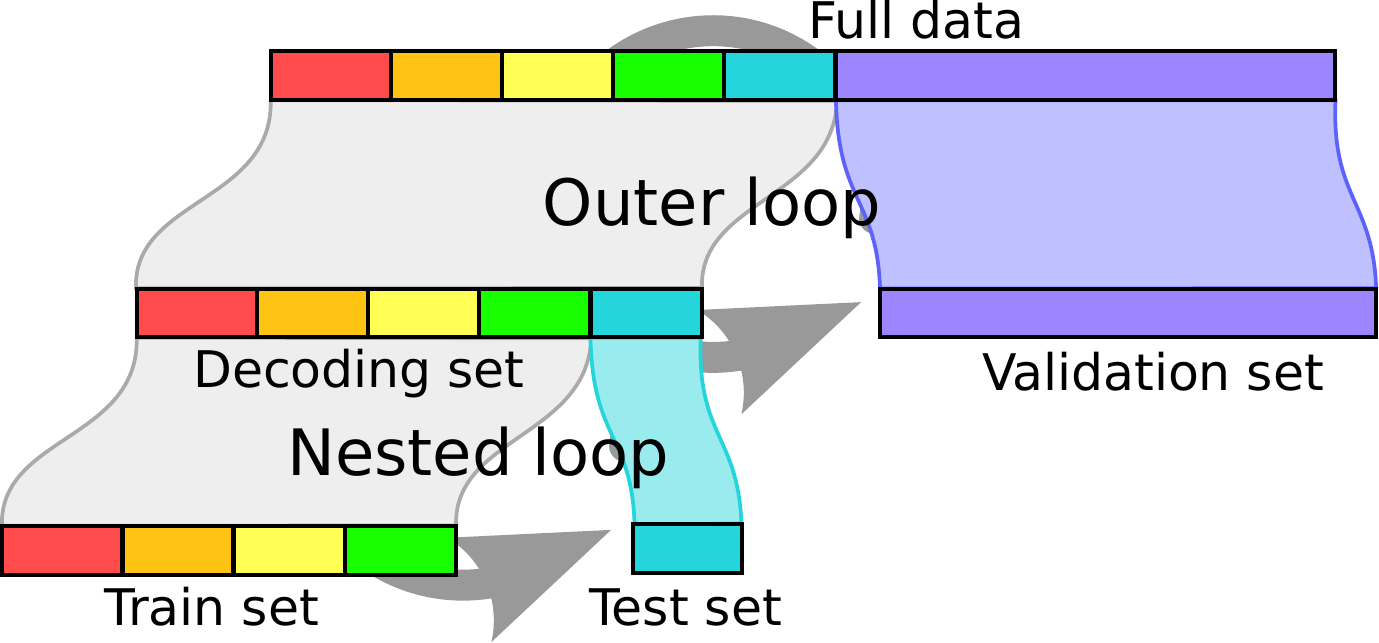}
\end{minipage}
\end{figure}

\paragraph{Nested cross-validation}

Choosing the right amount of regularization can improve the predictive power
of a decoder and
controls the appearance of the weight maps. The most common approach to set
it is to use cross-validation to measure predictive power for various
choices of regularization and to retain the value that maximizes
predictive power.
Importantly, with such a procedure, the amount of regularization becomes
a parameter adjusted on
data, and thus the predictive performance measured in the corresponding
cross-validation loop is not a reliable assessment of the predictive
performance of the model.
The standard procedure is then to refit the
model on the available data, and test its predictive performance on new
data, called a \emph{validation} set. Given a finite amount of data, a
\emph{nested cross-validation} procedure can be employed: the data are
repeatedly split in a validation set and a decoding set to perform
decoding. The decoding set itself is split in multiple train and test sets
with the same validation set, forming an inner ``nested'' cross-validation
loop used to set the regularization hyper-parameter, while the external loop varying
the validation set is used to measure prediction performance --see figure
\ref{fig:nested_cross_val}.

\paragraph{Model averaging} Choosing the best model in a
family of good models is hard. One option is to average the predictions
of a set of suitable models \cite[chap.\,35]{penny2007},
\cite{kuncheva2010classifier,churchill2014comparing,hoyos2015} --see
\cite[chap.\,8]{hastie2009elements} for a description outside of
neuroimaging. A simple version of this idea is
\emph{bagging} \cite{breiman1996bagging}: using \emph{bootstrap},
random resamplings of the data, to generate many train sets and
corresponding models, the predictions of which are then averaged.
The benefit of these approaches is that if the errors of each model are
sufficiently independent, they average out: the average model
performs better and displays much less variance.
With linear models often used as decoders in neuroimaging, model
averaging is appealing as it boils down to averaging weight maps.

To benefit from the stabilizing effect of model averaging in parameter
tuning, we can use a variant of both cross-validation and model
averaging\footnote{The combination of cross-validation and model
averaging is not new (see \emph{eg}
\cite{hoyos2015}), but it is seldom discussed in the neuroimaging
literature. It is commonly used in other areas of machine learning,
for instance to set parameters in bagged
models such as trees, by monitoring the out-of-bag error (\emph{eg} in
the \emph{scikit-learn} library \cite{pedregosa2011}).}. In a standard
cross-validation procedure, we repeatedly split the data in train and
test set and for each split, compute the test error for a grid of
hyper-parameter values. However, instead of selecting the hyper-parameter
value that minimizes the mean test error across the different splits, we
select \emph{for each split} the model that minimizes the corresponding
test error and average these models across splits.

\label{sec:averaging}%




\subsection{Model selection for neuroimaging decoders}


Decoding in neuroimaging faces specific model-selection challenges. The
main challenge is probably the scarcity of data relative to their dimensionality, typically hundreds of
observations\footnote{While in imaging neuroscience, hundreds of
observations seems
acceptably large, it is markedly below common sample sizes in machine
learning. Indeed, data analysis in brain imaging has historically been
driven by very simple models while machine learning has tackled
rich models since its inception.}. Another important aspect of decoding
is that, beyond predictive power, interpreting model weights is relevant.

\paragraph{Common decoders and their regularization}
Both to prefer simpler models and to facilitate interpretation, linear
models are ubiquitous in decoding. In fact, their weights form the common brain maps for visual interpretation.

\begin{figure}[b]
    \hfill%
    \includegraphics[width=.34\linewidth]{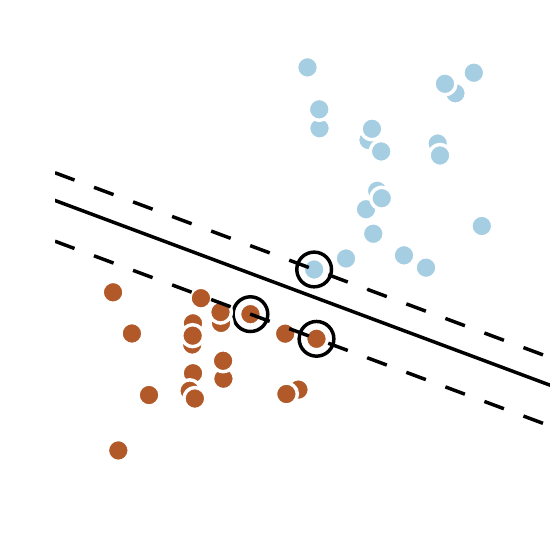}%
    \llap{\sffamily Large $C$\qquad}%
    \hfill%
    \includegraphics[width=.34\linewidth]{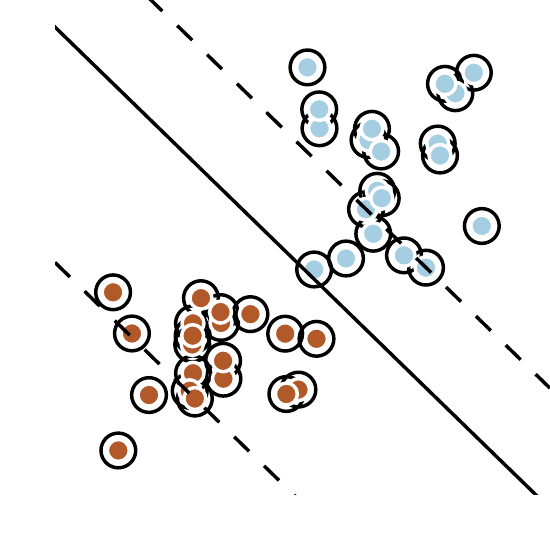}%
    \llap{\sffamily Small $C$\qquad}%
    \hfill\vbox{}%
   \caption{\textbf{Regularization with SVM-$\ell_2$}:
    blue and brown points are training samples of each class. The SVM
    learns a separating line between the two classes. In a weakly
    regularized setting (large $C$, this line is supported by few
    observations --called support vectors--, circled in black on the figure, while in a
    strongly-regularized case (small $C$), it is
    supported by the whole data.
    \label{fig:svm_reg}}%
\end{figure}

\begin{figure*}[bt]
\hspace*{-.025\linewidth}%
\begin{minipage}{.2\linewidth}
    \includegraphics[width=\linewidth]{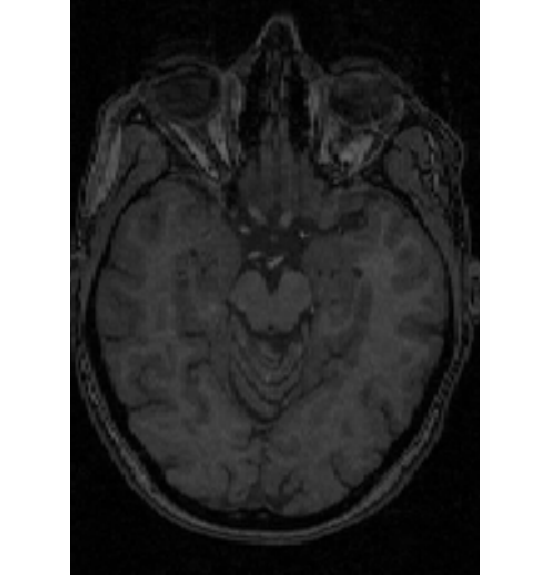}%
    \llap{\makebox[.65\linewidth][l]{%
	\raisebox{.01\linewidth}{\color{white}\small C=$10^{-5}$}}}%
\end{minipage}%
\hfill
\hspace*{-.075\linewidth}%
\begin{minipage}{.2\linewidth}
    \includegraphics[width=\linewidth]{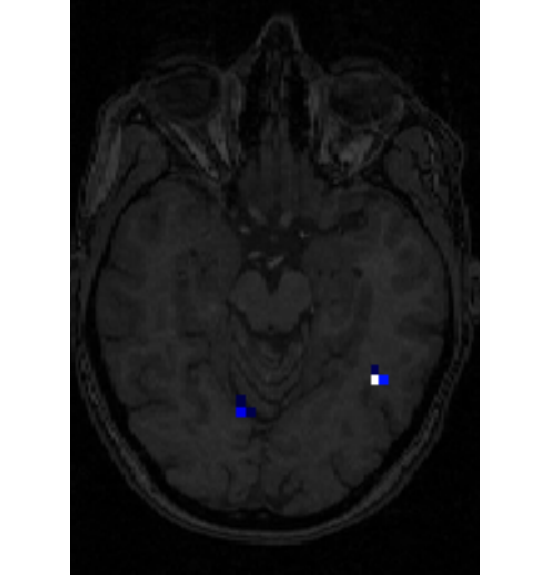}%
    \llap{\makebox[.65\linewidth][l]{%
	\raisebox{.01\linewidth}{\color{white}\small C=1.}}}%
\end{minipage}%
\hfill
\hspace*{-.075\linewidth}%
\begin{minipage}{.2\linewidth}
    \includegraphics[width=\linewidth]{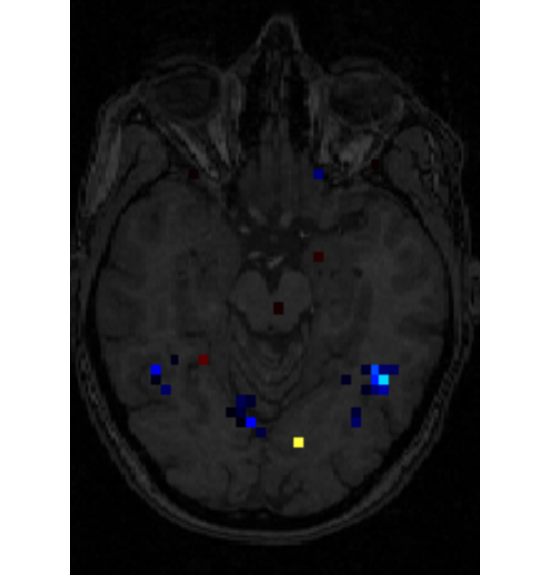}%
    \llap{\makebox[1.5\linewidth][c]{%
	\raisebox{.9\linewidth}{\color{white}\sffamily\bfseries Log-reg $\mathbf{\ell_1}$}}}%
    \llap{\makebox[.65\linewidth][l]{%
	\raisebox{.01\linewidth}{\color{white}\small C=$10^2$}}}%
\end{minipage}%
\hfill
\hspace*{-.075\linewidth}%
\begin{minipage}{.2\linewidth}
    \includegraphics[width=\linewidth]{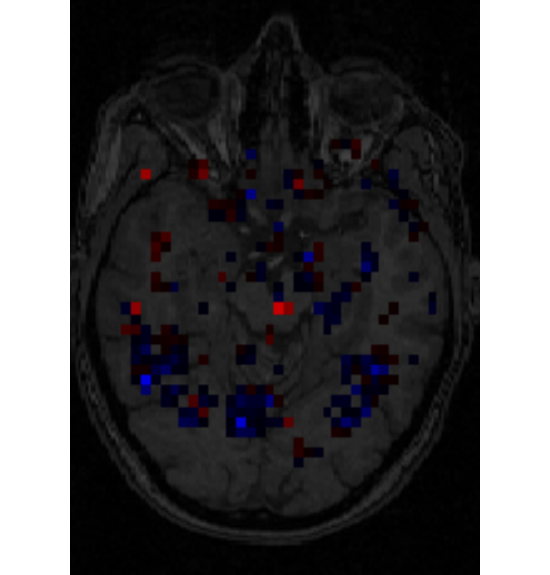}%
    \llap{\makebox[.65\linewidth][l]{%
	\raisebox{.01\linewidth}{\color{white}\small C=$10^{5}$}}}%
\end{minipage}%
\hfill%
\begin{minipage}{.2\linewidth}
    \includegraphics[width=\linewidth]{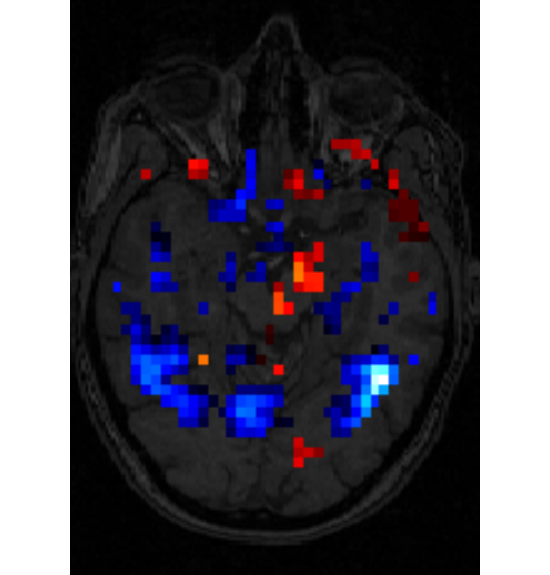}%
    \llap{\makebox[.65\linewidth][l]{%
	\raisebox{.01\linewidth}{\color{white}\small C=$10^{-5}$}}}%
\end{minipage}%
\hfill
\hspace*{-.075\linewidth}%
\begin{minipage}{.2\linewidth}
    \includegraphics[width=\linewidth]{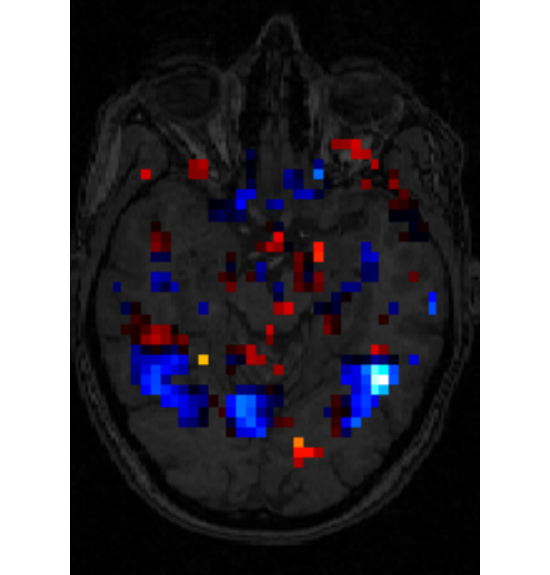}%
    \llap{\makebox[\linewidth][c]{%
	\raisebox{.9\linewidth}{\color{white}\sffamily\bfseries SVM $\mathbf{\ell_2}$}}}%
    \llap{\makebox[.65\linewidth][l]{%
	\raisebox{.01\linewidth}{\color{white}\small C=1}}}%
\end{minipage}%
\hfill
\hspace*{-.075\linewidth}%
\begin{minipage}{.2\linewidth}
    \includegraphics[width=\linewidth]{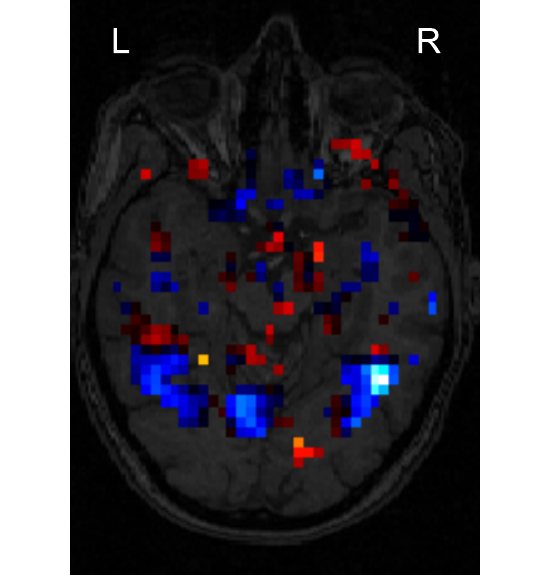}%
    \llap{\makebox[.65\linewidth][l]{%
	\raisebox{.01\linewidth}{\color{white}\small C=$10^{5}$}}}%
\end{minipage}%
\hspace*{-.025\linewidth}%

\caption{\textbf{Varying amount of regularization} on the face vs
house discrimination in the Haxby 2001 data \cite{haxby2001}.
\textbf{Left}: with a log-reg $\ell_1$, more regularization (small C)
induces sparsity. \textbf{Right}: with an SVM $\ell_2$, small C means that
weight maps are a combination of a larger number
of original images, although this has only a small visual impact on the
corresponding brain maps. \label{fig:maps}}
\end{figure*}

The classifier used most often in fMRI is the support vector machine
(SVM) \cite{mouraomiranda2005,chen2006exploring,laconte2005support}.
However, logistic
regressions (Log-Reg) are also often used
\cite{ryali2010,varoquaux2012icml,chen2006exploring,yamashita2008,rasmussen2012model}. Both of these
classifiers learn a linear model by minimizing the sum of a \emph{loss}
$\mathcal{L}$ --a data-fit term-- and a \emph{penalty} $p$
--the regularizing energy term that favors simpler models:
\begin{equation*}
    \hat{\mathbf{w}} = \underset{\mathbf{w}}{\text{argmin}}\bigl(
	\mathcal{L}(\mathbf{w}) + \frac{1}{C} \, p(\mathbf{w})
    \bigr)
    \qquad
    \smash{\mathcal{L} =
    \begin{cases}
	\text{\raisebox{-1ex}{%
	    \includegraphics[height=3ex]{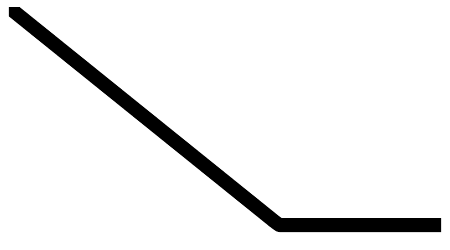}}%
	}
	& \hspace*{-1.5ex}\text{SVM} \\
	\text{\raisebox{-1ex}{%
	    \includegraphics[height=3ex]{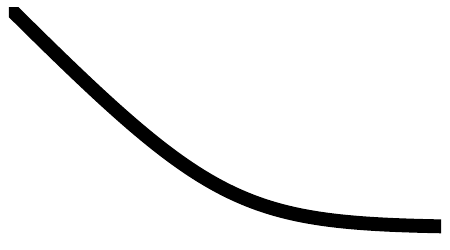}}%
	}
	& \hspace*{-1.5ex}\text{logistic}
    \end{cases}}
\end{equation*}
where $C$ is the regularization parameter that controls the
bias-variance tradeoff: small $C$ means strong regularization.
The SVM and logistic regression model differ only by the loss used. For
the SVM the loss is a hinge loss: flat and exactly zero for
well-classified samples and with a misclassification cost increasing linearly
with distance to the decision boundary. For the logistic regression, it
is a logistic loss, which is a soft, exponentially-decreasing, version of
the hinge \cite{hastie2009elements}.
By far the most common regularization is the $\ell_2$ penalty. Indeed, the common form of SVM uses $\ell_2$
regularization, which we will denote SVM-$\ell_2$. Combined with the
large zero region of the hinge loss, strong $\ell_2$ penalty implies that
SVMs build
their decision functions by combining a small number of training images
(see figure \ref{fig:svm_reg}).
Logistic regression is similar: the loss has no flat region, and thus
every sample is used, but some very weakly.
Another frequent form of penalty, $\ell_1$, imposes sparsity on the
weights: a strong regularization means that the weight maps $\mathbf{w}$
are mostly comprised of zero voxels (see Fig.\,\ref{fig:maps}).


\paragraph{Parameter-tuning in neuroimaging}

In neuroimaging, many publications do not discuss their choice of decoder
hyper-parameters; while others state that they use the default value,
\emph{eg} $C = 1$ for SVMs. Standard machine learning practice advocates
setting the parameters by nested cross-validation
\cite{hastie2009elements}. For non sparse, $\ell_2$-penalized models, the
amount of regularization often does not have a strong influence on the
weight maps of the decoder (see figure \ref{fig:maps}). Indeed,
regularization in these models changes the fraction of input maps
supporting the hyperplane (see \ref{fig:svm_reg}). As activation maps
for the same condition often have similar aspects,
this fraction impacts weakly decoders' maps.

For sparse models, using the $\ell_1$ penalty, sparsity is often seen as a
means to select relevant voxels for prediction
\cite{carroll2009,ryali2010}. In this case, the amount of regularization
has a very visible consequence on weight maps and voxel
selection (see figure \ref{fig:maps}). Neuroimaging studies often
set it by cross-validation \cite{carroll2009}, though very seldom nested
(exceptions comprise \cite{churchill2014comparing,varoquaux2012icml}).
Voxel selection by $\ell_1$ penalty on brain maps is unstable because
neighboring voxels respond similarly
 and $\ell_1$ estimators will choose somewhat randomly few of these correlated features \cite{varoquaux2012icml,rondina2013stability}.
Hence various strategies combining sparse models are used in
neuroimaging to improve decoding performance and stability.
Averaging weight maps across cross-validation folds
\cite{hoyos2015,varoquaux2012icml}, as described above,
is interesting, as it stays in the realm of linear models.
Relatedly, \cite{grosenick2013} report the median of weight maps,
thought it does not correspond to weights in a predictive model.
Consensus between sparse models over data
perturbations gives theoretically better feature selection 
\cite{meinshausen2010stability}. In fMRI, it has been used to screen
voxels before fitting linear models
\cite{rondina2013stability,varoquaux2012icml} or to interpret selected
voxels \cite{yamashita2008}.


For model selection in neuroimaging, 
prediction performance is not the only relevant metric and
some control over the estimated model weights is also important. For
this purpose,
\cite{laconte2003,strother2002quantitative,rasmussen2012model} advocate using a tradeoff between
prediction performance and stability of decoder maps. Stability is a
proxy for estimation error on these maps, a quantity that is not
accessible without knowing the ground truth. While very useful it
gives only indirect information on estimation error: it does not control
whether all the predictive brain regions were found, nor whether all
regions found are predictive. Indeed, a decoder choosing its maps
independently from the data would be very stable, though likely with poor
prediction performance. Hence the challenge is in
finding a good prediction-stability tradeoff \cite{strother2014stability,
rasmussen2012model}.
%


\section{Empirical studies: cross-validation at work}

Here we highlight practical aspects of
cross-validation in brain decoding with simple experiments. We first demonstrate the variability
of prediction estimates on MRI, MEG, and simulated data. 
We then explore how to tune decoders parameters.

\subsection{Experiments on real neuroimaging data}

\paragraph{A variety of decoding datasets}
To achieve reliable empirical conclusions, it is
important to consider a large number of different neuroimaging
studies. We investigate cross-validation in a
large number of 2-class classification problems, from 7 different
fMRI datasets (an exhaustive list can be found in 
\ref{sec:table}). We decode visual stimuli within subject (across sessions)
in the classic Haxby dataset \cite{haxby2001}, and across subjects using data
from \cite{duncan2009consistency}.
We discriminate across subjects \emph{i)} affective content with data
from \cite{wager2008neural}, \emph{ii)} visual from narrative with data from
\cite{moran2012social}, \emph{iii)} famous, familiar, and scrambled faces from a visual-presentations dataset
\cite{henson2002face}, and \emph{iv)} left and right saccades in data from
\cite{knops2009recruitment}.
We also use a
non-published dataset, ds009 from openfMRI \cite{poldrack2013openfmri}.
All the across-subject predictions are performed on trial-by-trial response
(Z-score maps) computed in a first-level GLM. Finally, beyond fMRI, we perform
prediction of gender from VBM maps using the OASIS data \cite{marcus2007}.
Note that all these
tasks cover many different
settings, range from easy discriminations to hard ones, and
(regarding fMRI) recruit very different systems with different effect
size and variability. The number of observations available to the
decoder varies between 80 (40 per class) and 400, with balanced classes.

The results and figures reported below are for all these datasets.
We use more inter-subject than intra-subject datasets. However 15
classification tasks out of 31 are intra-subject (see
Tab.\,\ref{tab:results}). In addition, when decoding is performed
intra-subject, each subject gives rise to a cross-validation.
Thus in our cross-validation study,
82\% of the data points are for intra-subject settings.

All MR data but \cite{knops2009recruitment}
are openly available from openfMRI
\cite{poldrack2013openfmri} or OASIS \cite{marcus2007}.
Standard preprocessing and first-level analysis were applied using
SPM, Nipype and Nipy (details in \ref{sec:fmri_data_analysis}).  All
MR data were variance-normalized\footnote{Division of each time series
  voxel/MEG sensor by its standard deviation} and spatially-smoothed
at 6\,mm FWHM for fMRI data and 2\,mm FWHM for VBM data.

\paragraph{MEG data}
Beyond MR data, we assess cross-validation strategies for decoding of
event-related dynamics in neurophysiological data. We analyze
magneteoencephalography (MEG) data from a working-memory experiment made available by the Human Connectome Project~\cite{larson2013adding}.
We perform intra-subject decoding in 52 subjects with two runs, using a
temporal window on the sensor signals (as in
\cite{sitt2014large}). Here, each run serves as validation
set for the other run. We consider two-class decoding problems, focusing on either the image content (faces vs tools) or the
functional role in the working memory task (target vs low-level and
high-level distractors). This yields in total four classification analyzes
per subject. For each trial, the feature set is a time window constrained
to 50\,ms before and 300\,ms after event onset, emphasizing visual components.
We use the cleaned single-trial outputs from
the HCP ``tmegpreproc'' pipeline. MEG data analysis was performed with the
MNE-Python software~\cite{gramfort2013meg, gramfort2014mne}. Full details
on the analysis are given in \ref{sec:meg_data_analysis}.

\paragraph{Experimental setup}
Our experiments make use of nested cross-validation for an accurate
measure of prediction power. As in figure \ref{fig:nested_cross_val}, we
repeatedly split the data in a validation set and a decoding set passed on to the
decoding procedure (including parameter-tuning for experiments in 
\ref{sec:cv_for_tuning} and \ref{sec:stability}). To get a good measure of
predictive power, we choose large validation sets of 50\% of
the data, respecting the sample dependence structure (leaving out
subjects, or sessions). We use 10 different validation sets that each
contribute a data point in results.

We follow standard decoding practice in fMRI \cite{pereira2009machine}.
We use univariate feature selection on the training set to select the
strongest 20\% of voxels and train a decoder
on the selected features. As a choice of decoder, we explore classic linear models: SVM and
logistic regression with $\ell_1$ and $\ell_2$ penalty\footnote{Similar
decoders
adding a regularization that
captures spatio-temporal correlations among the voxels are
well suited for neuroimaging
\cite{gramfort2013,michel2011tv,grosenick2013,karahanouglu2013totalactivation}.
Also, random forests, based on model averaging discussed above,
have been used in fMRI
\cite{langs2011gini,kuncheva2010classifier}.
However, this review focuses on the most common practice. 
Indeed, these decoders entail computational costs 
that are intractable given the number of models fit in the
experiments.
}.
We use scikit-learn for all decoders \cite{pedregosa2011,abraham2014}.

In a first experiment, we compare decoder performance estimated by
cross-validation on the decoding set, with performance measured on the
validation set. In a second experiment, we investigate the use of
cross-validation to tune the model's regularization parameter, either
using the standard \emph{refitting} approach, or \emph{averaging} as
described in section \ref{sec:averaging}, as well as using the default
$C=1$ choice of parameter, and a value of $C = 1000$.

\subsection{Results: cross-validation to assess predictive power}

\begin{figure}[b]
\centerline{%
\includegraphics[height=.65\linewidth,
		 trim={0 .4cm 0 0}]{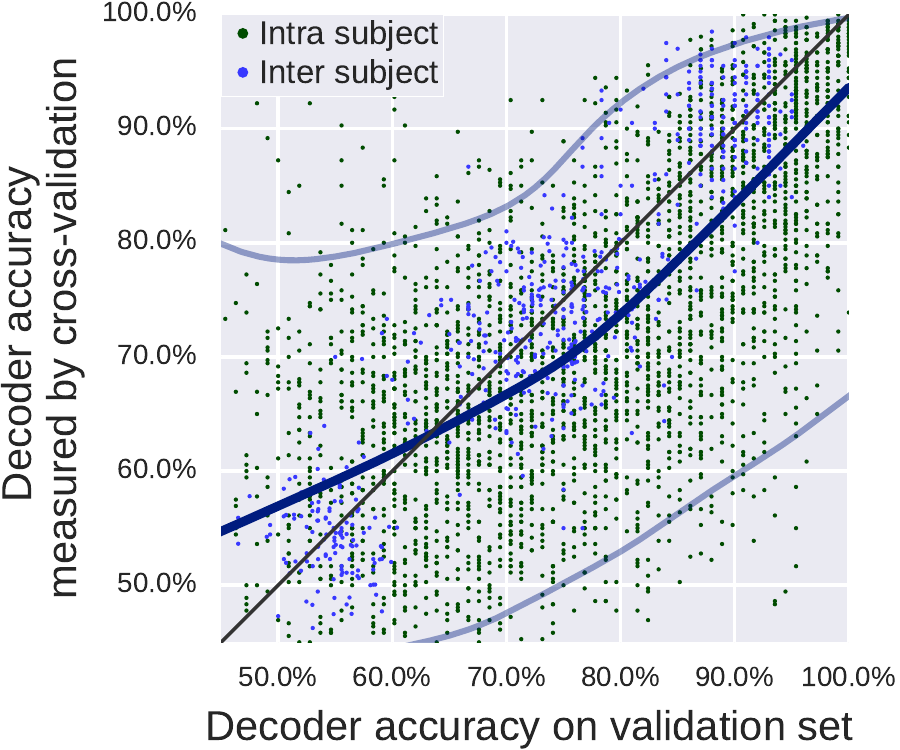}%
}%

\caption{\textbf{Prediction error: cross-validated versus validation
set}. Each point is a measure of predictor error in the inner
cross-validation loop (10 splits, leaving out 20\%), or in
the left-out validation set.
The dark line is an indication of the tendency, using a lowess 
local regression. The two light lines indicate the boundaries above and
below which fall 5\% of the points.
They are estimated using a Gaussian kernel
density estimator, with a bandwidth set by the Scott method, and
computing the CDF along the y direction.
\label{fig:cv_error}}
\end{figure}

\paragraph{Reliability of the cross-validation measure}
Considering that prediction error on the large left-out validation set
is a good estimate of predictive power, we use it to assess the
quality of the estimate given by the nested cross-validation loop.
Figure \ref{fig:cv_error} shows the prediction error measured by
cross-validation as a function of the validation-set error across all
datasets and validation splits. It reveals a small negative bias: as
predicted by the theory, cross-validation is pessimistic compared to a
model fit on the complete decoding set. However,
models that perform poorly are often reported with a better performance by
cross-validation. Additionally, cross-validation estimates display a large
variance: there is a scatter between estimates in the nested
cross-validation loop and in the validation set.

\begin{figure}\smallskip
{\small\sffamily
Cross-validation\hfill {\bfseries Difference in accuracy
measured}\hfill\vbox{}\vspace*{-.4ex}

strategy\hfill\hfill {\bfseries ~~by cross-validation and on validation set}\hfill\vbox{}}
\includegraphics[width=\linewidth]{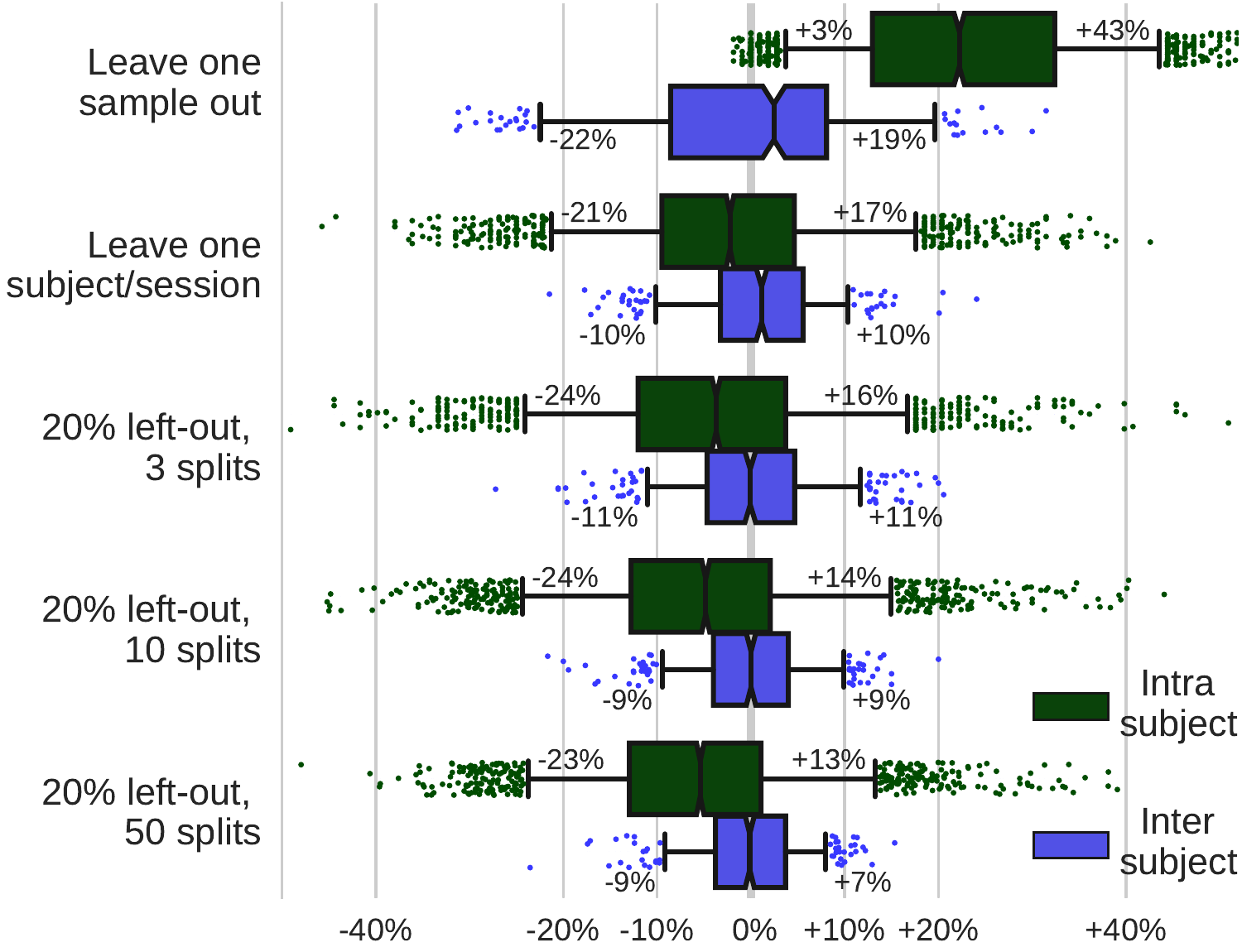}\\[-.55em]
{\sffamily\tiny\hspace*{.195\linewidth}\scalebox{1.15}{cross-validation
$<$ validation set} \hfill
\scalebox{1.15}{cross-validation $>$ validation set}}

\caption{\textbf{Cross-validation error: different strategies}.
    Difference between accuracy measured by cross-validation and on the
    validation set, in intra and inter-subject settings, for different
    cross-validation strategies: leave one sample out,
    leave one block of samples out (where the block is the natural
    unit of the experiment: subject or session), or
    random splits leaving out 20\% of the blocks as test data, with 3, 10, or
    50 random splits.
    For inter-subject settings, leave one sample out corresponds to
    leaving a session out.
    The box gives the quartiles, while the whiskers give
    the 5 and 95 percentiles.\label{fig:cv_strategies}}
\end{figure}

\paragraph{Different cross-validation strategies}
Figure \ref{fig:cv_strategies} summarizes the discrepancy between
prediction accuracy measured by cross validation and on the validation
set for different cross-validation strategies: leaving one sample out,
leaving one block of data out --where blocks are the natural
units of the experiment, sessions or subjects-- and random splits leaving
out 20\% of the blocks of data with 3, 10, and 50 repetitions.

Ideally, a good cross-validation strategy would minimize this
discrepancy. We find that leave-one-sample out is very optimistic in
within-subject settings. This is expected, as samples are highly
correlated. When leaving out blocks of data that minimize dependency
between train and test set, the bias mostly disappears. The remaining
discrepancy appears mostly as variance in the estimates of prediction
accuracy. For repeated random splits of the data, the larger the number
of splits, the smaller the variance. Performing 10 to 50 splits with 20\%
of the data blocks left out gives a better estimation than leaving
successively each blocks out, at a fraction of the computing cost if the
number of blocks is large.
While intra and inter subject settings do not differ strongly when
leaving out blocks of data, intra-subject settings display a
larger variance of estimation as well as a slight negative bias. These
are likely due to non-stationarity in the time-series, \emph{eg} scanner
drift or loss of vigilance. In inter-subject settings, heterogeneities
may hinder prediction \cite{strother2014stability}, yet a
cross-validation strategy with multiple subjects in the test set will
yield a good estimate of prediction accuracy\footnote{The probability of 
correct classification for each subject is also an interesting quantity, though 
it is not the same thing as the prediction accuracy measured by
cross-validation \cite[sec 7.12]{hastie2009elements}. It can be computed by non-parametric approaches such as
bootstrapping the train set \cite{platt1999probabilistic}, or using a
posterior probability, as given by certain classifiers.}.

\paragraph{Other modalities: MEG and simulations}
We run the experiments on the MEG decoding tasks and the simulations.

We generate simple simulated data that mimic brain imaging to better understand trends and limitations of cross-validation. Briefly,
we generate data with 2 classes in 100 dimensions with Gaussian noise
temporally auto-correlated and varying the separation between the class
centers (more details in \ref{sec:simulated_data}). We run the
experiments on a decoding set of 200 samples.

\begin{figure}\smallskip
{\small\sffamily
Cross-validation\hfill {\bfseries Difference in accuracy
measured}\hfill\vbox{}\vspace*{-.4ex}

strategy\hfill\hfill {\bfseries ~~by cross-validation and on validation set}\hfill\vbox{}}
\includegraphics[width=\linewidth]{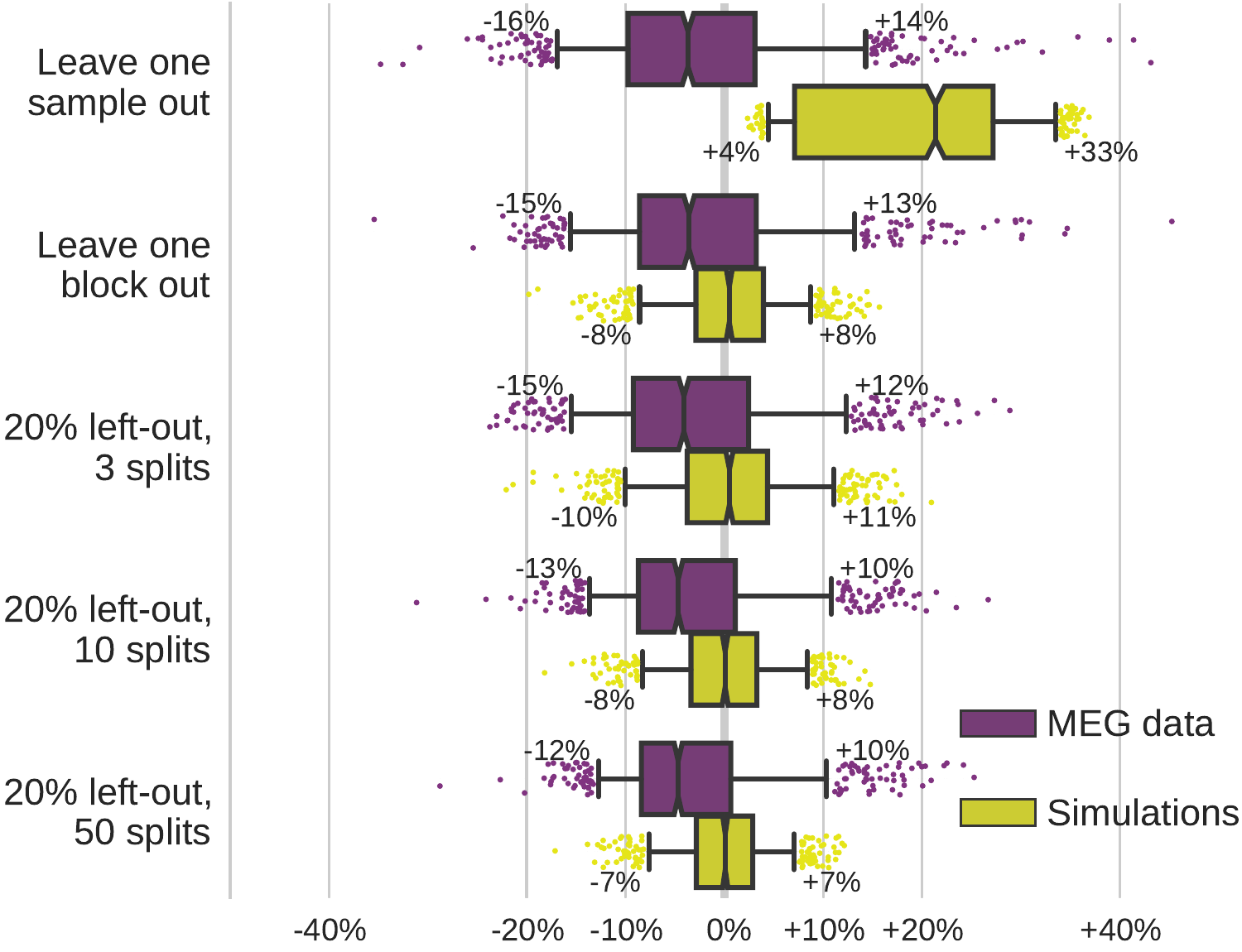}\\[-.55em]
{\sffamily\tiny\hspace*{.195\linewidth}\scalebox{1.15}{cross-validation
$<$ validation set} \hfill
\scalebox{1.15}{cross-validation $>$ validation set}}

\caption{\textbf{Cross-validation error: non-MRI modalities}.
    Difference between accuracy measured by cross-validation and on the
    validation set, for MEG data and simulated data, with different
    cross-validation strategies.
    Detailed simulation results in \ref{sec:simulated_data}.
    \label{fig:cv_strategies_other_data}}
\end{figure}

The results, displayed in figure \ref{fig:cv_strategies_other_data},
reproduce the trends observed on MR data. As the simulated data is temporally
auto-correlated, leave-one-sample-out is strongly optimistic. Detailed
analysis varying the separability of the classes 
(\ref{sec:simulated_data}) shows that cross-validation tends to be
pessimistic for high-accuracy situations, but optimistic when prediction
is low. For MEG
decoding, the leave-one-out procedure is on trials, and thus does not
suffer from  correlations between samples.
Cross-validation is slightly pessimistic and display a large
variance, most likely because of inhomogeneities across samples. In both
situations, leaving blocks of data out with many splits
(\emph{e.g.} 50) gives best results.

\begin{figure*}
\hspace*{-.01\linewidth}%
\begin{minipage}{.29\linewidth}
    \includegraphics[width=\linewidth]{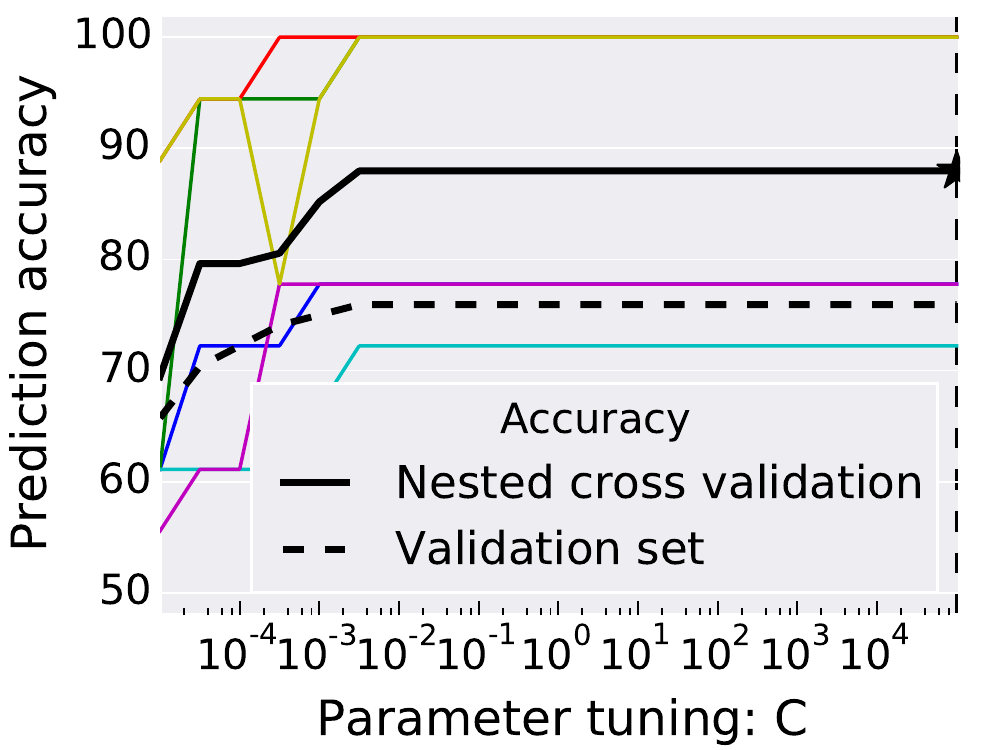}%
    \llap{\makebox[.7\linewidth][c]{%
	\raisebox{.64\linewidth}{\small\sffamily SVM $\ell_2$}}}
\end{minipage}%
\hfill
\hspace*{-.06\linewidth}%
\begin{minipage}{.29\linewidth}
    \includegraphics[width=\linewidth]{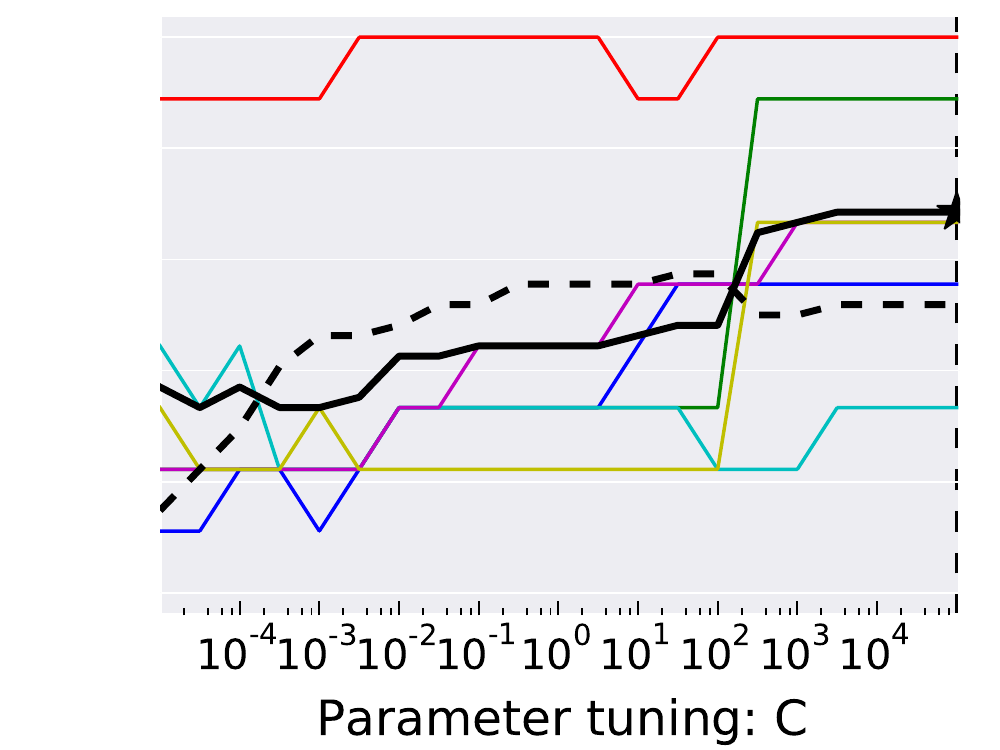}%
    \llap{\makebox[.65\linewidth][l]{%
	\raisebox{.64\linewidth}{\small\sffamily Log-reg $\ell_2$}}}
\end{minipage}%
\hfill
\hspace*{-.06\linewidth}%
\begin{minipage}{.29\linewidth}
    \includegraphics[width=\linewidth]{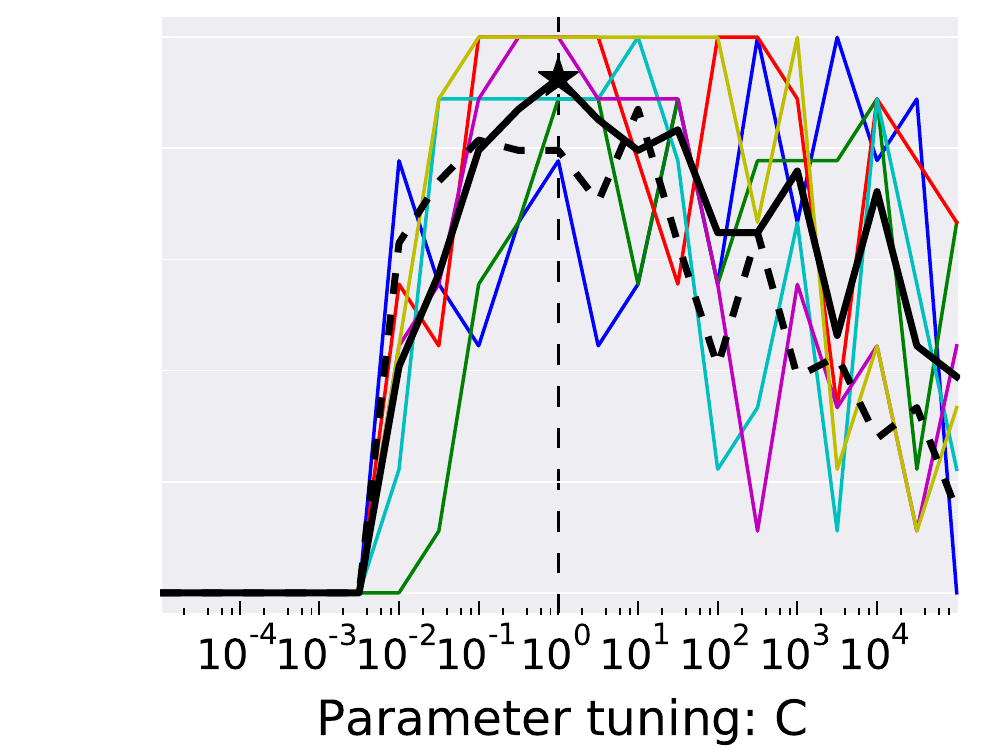}%
    \llap{\makebox[.8\linewidth][l]{%
	\raisebox{.64\linewidth}{\small\sffamily SVM $\ell_1$}}}
\end{minipage}%
\hfill
\hspace*{-.06\linewidth}%
\begin{minipage}{.29\linewidth}
    \includegraphics[width=\linewidth]{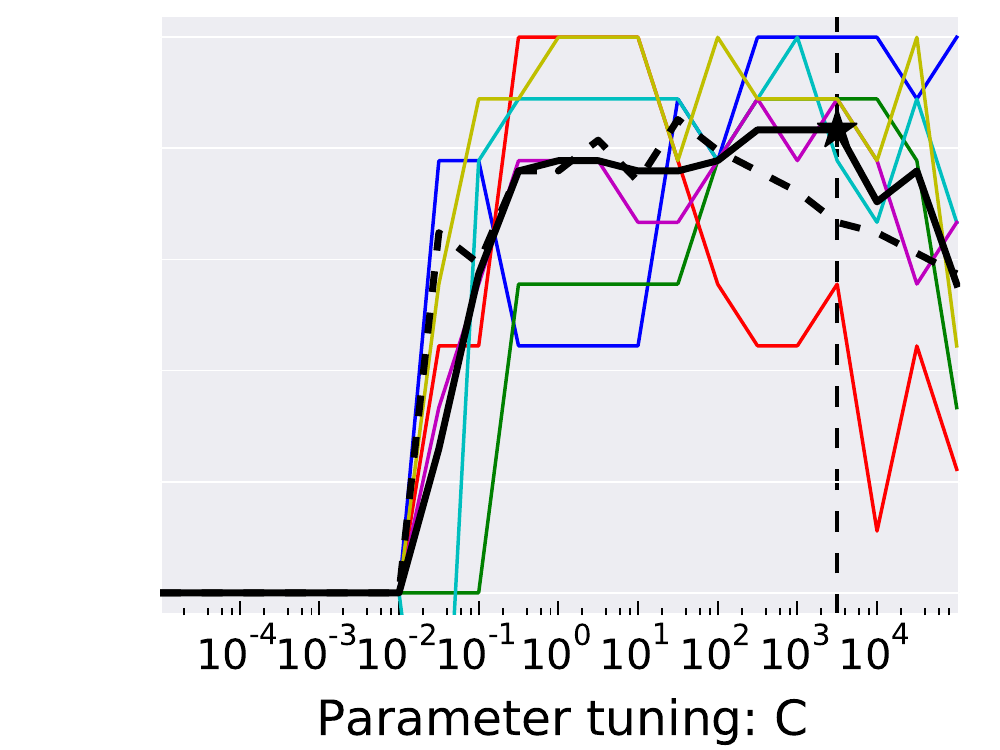}%
    \llap{\makebox[.82\linewidth][l]{%
	\raisebox{.64\linewidth}{\small\sffamily Log-reg $\ell_1$}}}
\end{minipage}%
\hfill
\caption{\textbf{Tuning curves} for SVM $\ell_2$, logistic
regression $\ell_2$, SVM $\ell_1$, and logistic
regression $\ell_1$, on the scissors / scramble discrimination for
the Haxby dataset \cite{haxby2001}. The thin colored lines are test
scores for each of the internal cross-validation folds, the thick
black line is the average of these test scores on all folds, and the
thick dashed line is the score on left-out validation data. The vertical
dashed line is the parameter selected on the inner cross-validation score.
\label{fig:tuning_curves}}
\end{figure*}

\subsection{Results on cross-validation for parameter tuning}

\label{sec:cv_for_tuning}

We now evaluate cross-validation as a way of setting decoder
hyperparameters.

\paragraph{Tuning curves: opening the black box}
Figure \ref{fig:tuning_curves} is a didactic view on the
parameter-selection problem: it gives, for varying values of the meta-parameter
$C$, the cross-validated error and the validation error for a given split
of validation data\footnote{For the figure, we compute cross-validated
error with a leave-one-session-out
on the first 6 sessions of the scissor / scramble Haxby data, and use the
last 6 sessions as a validation set.}. The validation error is computed on a large sample size
on left out data, hence it is a good estimate of the generalization error
of the decoder. Note that the parameter-tuning procedure does not have
access to this information. The discrepancy between the tuning curve,
computed with cross-validation on the data available to the decoder, and
the validation curve, is an indication of the uncertainty on the
cross-validated estimate of prediction power. Test-set error curves of
individual splits of the nested cross-validation loop show plateaus and a
discrete behavior. Indeed, each individual test set contains dozens
of observations. The small combinatorials limit the accuracy of error
estimates.

Figure \ref{fig:tuning_curves} also shows that non-sparse
--$\ell_2$-penalized-- models are not very sensitive to the choice of
the regularization parameter C: the tuning curves display a wide plateau\footnote{This plateau
is due to the flat,
or nearly flat, regions of their loss that renders them mostly dependent only
on whether samples are well classified or not.}. However, for
sparse models ($\ell_1$ models), the maximum of the tuning curve is a more
narrow peak, particularly so for SVM.
A narrow peak in a tuning curve implies that a choice of optimal parameter
may not
alway carry over to the validation set.

\begin{figure}[!b]
\setlength{\mylength}{20cm}%
\includegraphics[height=.238\mylength,
		 trim={0 .4cm 0 0}]{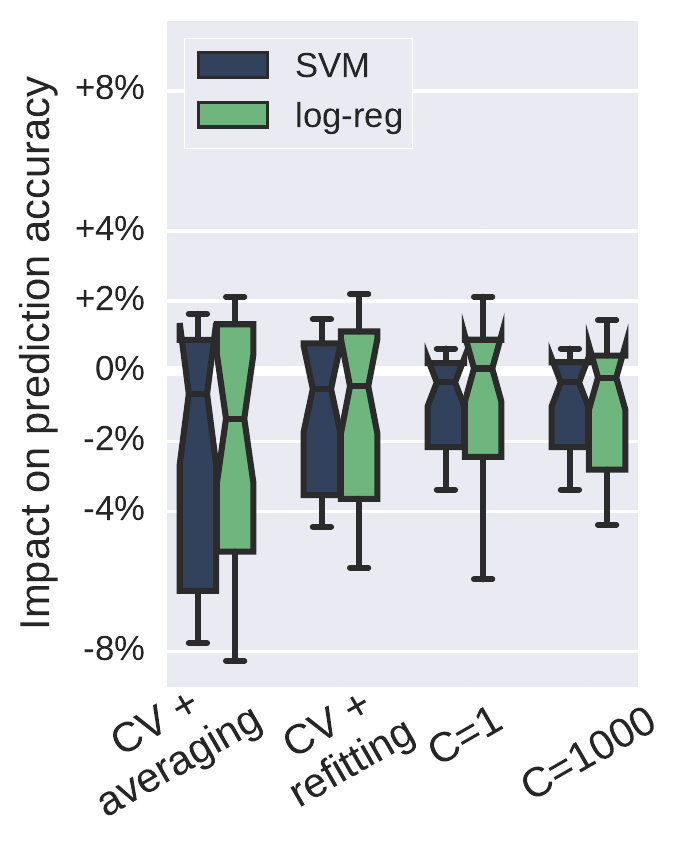}%
\llap{\raisebox{.232\mylength}{%
    \rlap{\small\bfseries\sffamily Non-sparse models}}%
    \hspace*{.34\linewidth}%
    }%
\hfill%
\includegraphics[height=.238\mylength,
		 trim={0 .4cm 0 0}]{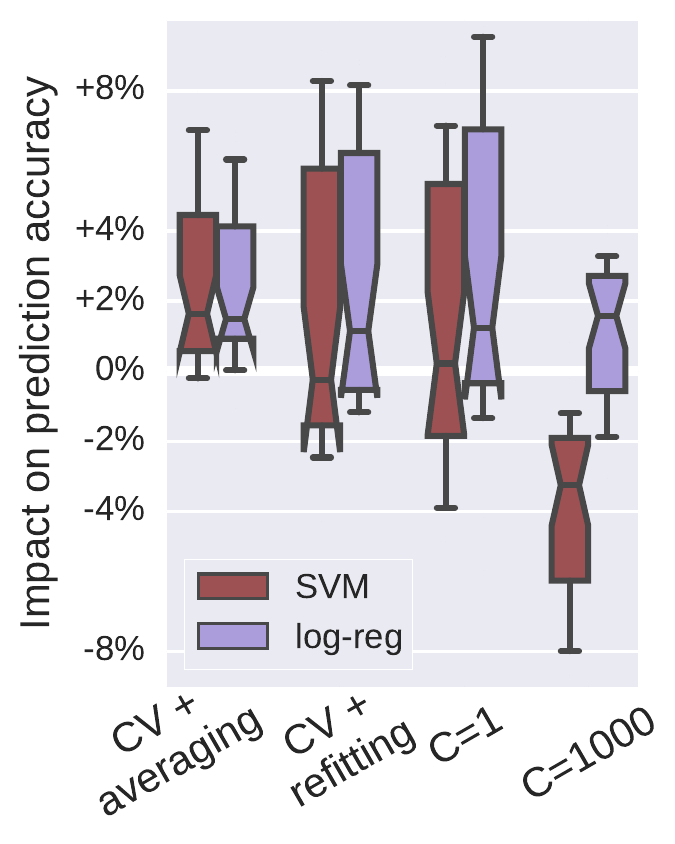}%
\llap{\raisebox{.232\mylength}{%
    \rlap{\small\bfseries\sffamily Sparse models}}%
    \hspace*{.34\linewidth}%
    }

\caption{\textbf{Prediction accuracy: impact of the parameter-tuning strategy}. For each strategy, difference to the mean prediction
accuracy in a given validation split.
\label{fig:parameter_tuning}}
\end{figure}

\paragraph{Impact of parameter tuning on prediction accuracy}
Cross-validation is often used to select regularization hyper-parameters,
\emph{eg} to control the amount of sparsity. On figure
\ref{fig:parameter_tuning}, we compare the various strategies: refitting
with the best parameters selected by nested cross-validation, averaging
the best models in the nested cross-validation, or simply using either
the default value of C or a large one, given that tuning curves can
plateau for large C.

For non-sparse models, the figure shows that tuning the hyper-parameter by
nested cross validation does not lead in general to better prediction
performance than a default choice of hyper-parameter. Detailed
investigations (figure
\ref{fig:parameter_tuning_ext}) show that these conclusions hold well
across all tasks, though refitting after nested cross-validation
is beneficial for good prediction accuracies, \emph{ie} when
there is either a large signal-to-noise ratio or many samples.

For sparse models, the picture is slightly different. Indeed,
high values of C lead to poor performance --presumably as the models are
overly sparse--, while using default value $C=1$, refitting or averaging
models tuned by cross-validation all perform well.
Investigating how these compromises vary as a function of model accuracy
(figure \ref{fig:parameter_tuning_ext})
reveals that for difficult decoding situations (low prediction) it is
preferable to use the default $C=1$, while in good decoding situations,
in is beneficial to tune C by nested cross-validation and rely on model
averaging, which tends to perform well and displays less variance.

\subsection{Results: stability of model weights}

\label{sec:stability}

\begin{figure}[!b]
\setlength{\mylength}{20cm}%
\includegraphics[height=.238\mylength,
		 trim={0 .4cm 0 0}]{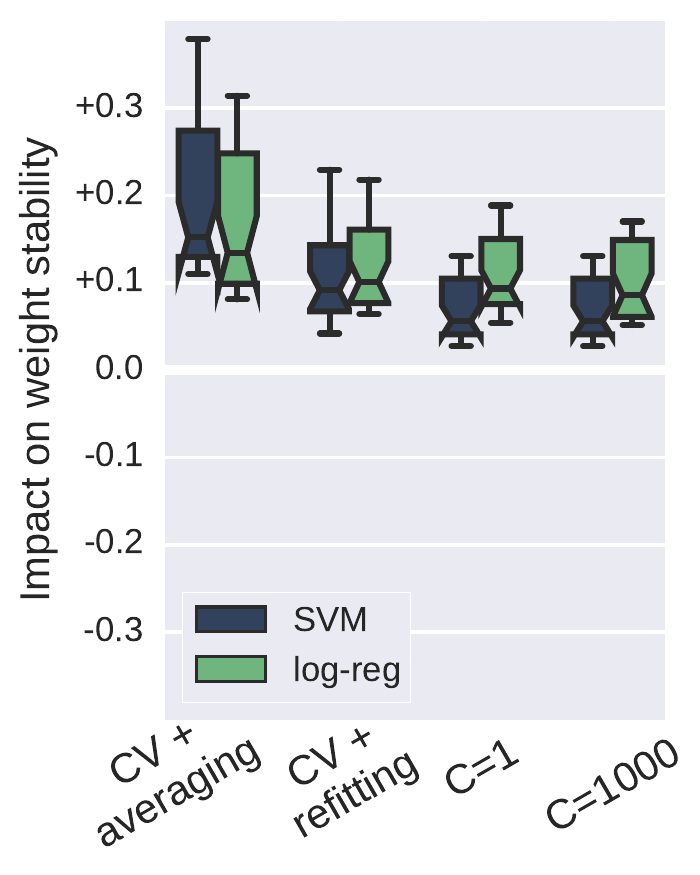}%
\llap{\raisebox{.234\mylength}{%
    \rlap{\small\bfseries\sffamily Non-sparse models}}%
    \hspace*{.34\linewidth}%
    }%
\hfill%
\includegraphics[height=.238\mylength,
		 trim={0 .4cm 0 0}]{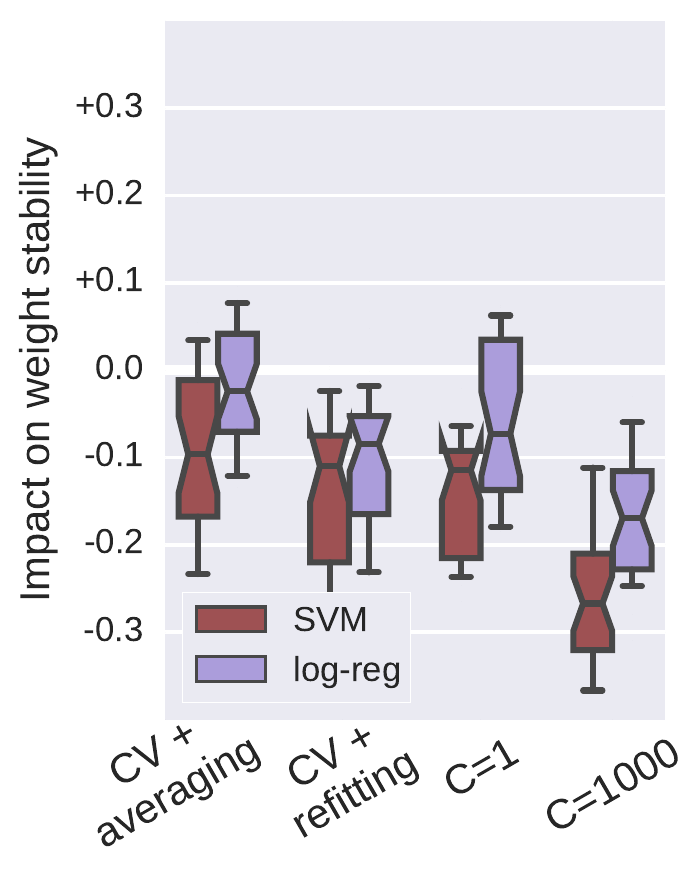}%
\llap{\raisebox{.234\mylength}{%
    \rlap{\small\bfseries\sffamily Sparse models}}%
    \hspace*{.34\linewidth}%
    }

\caption{\textbf{Stability of the weights: impact of the parameter-tuning
strategy}: for each strategy, difference to the mean stability of the
model weights, where the stability is the correlation of the weights
across validation splits. As the stability is a correlation, the unit is
a different between correlation values. The reference is the mean
stability across all models for a given prediction task.
\label{fig:stability}}
\end{figure}

\begin{figure}[!b]
\centerline{%
\includegraphics[width=.78\linewidth,
		 trim={0 0cm .1cm 0}]{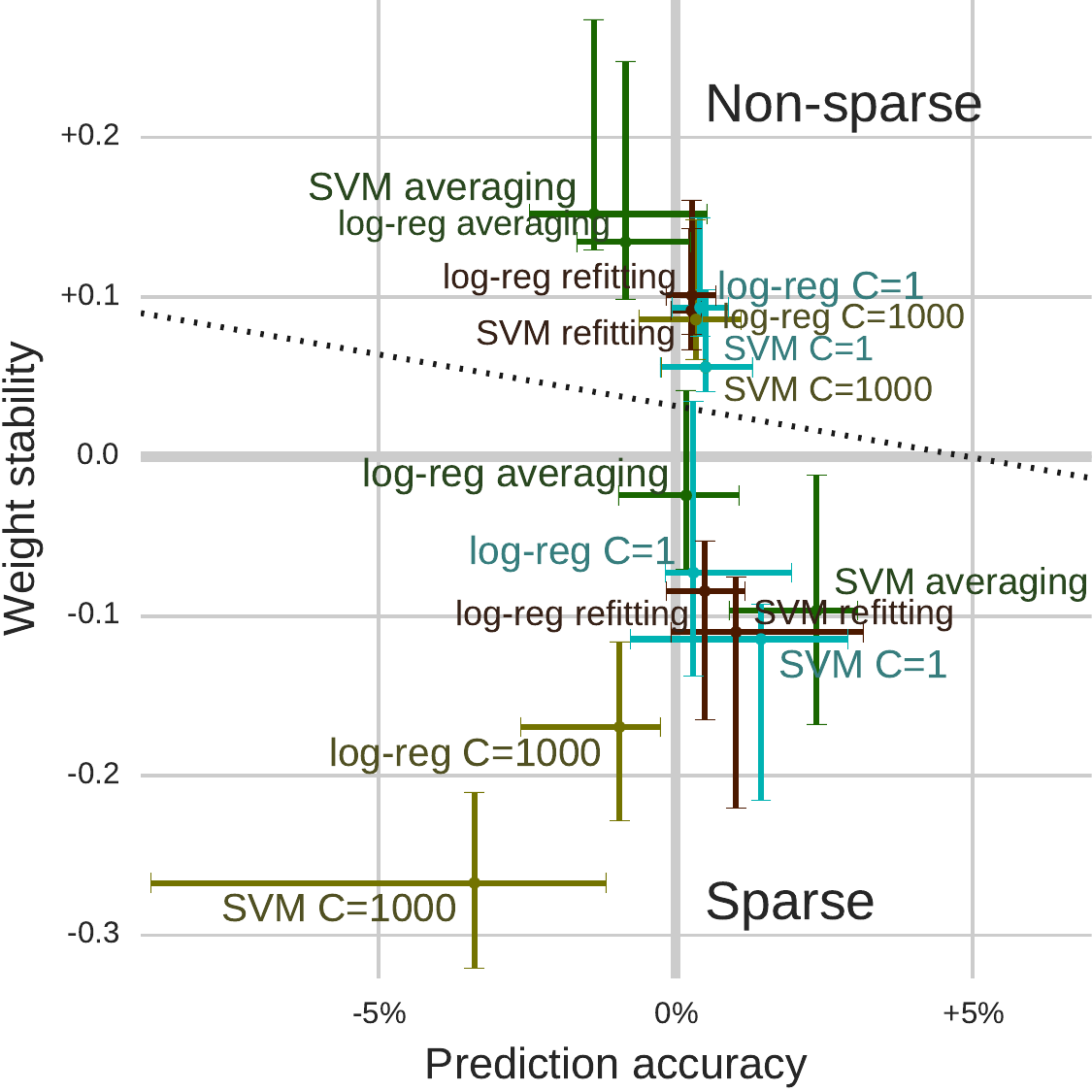}}%

\caption{\textbf{Prediction -- stability tradeoff}
The figure summarize figures \ref{fig:parameter_tuning} and
\ref{fig:stability}, reporting
the stability of the weights, relative to the split's average,
as a function of the delta in prediction accuracy.
It provides an overall summary, with 
errors bar giving the first and last quartiles (more detailed figures in 
\ref{sec:stability_prediction_appendix})
\label{fig:tradeoff}}
\end{figure}

\paragraph{Impact of parameter tuning on stability}
The choice of regularization parameter also affects the stability of the
weight maps of the classifier. Strongly regularized maps underfit, thus
depending less on the train data, which may lead to increased stability.
We measure stability of the decoder maps by computing their correlation
across different choices of validation split for a given task.

Figure \ref{fig:stability} summarizes the results on stability.
For all models, sparse and non-sparse, model averaging does
give more stable maps, followed by refitting after choosing parameters by
nested cross-validation.
%
Sparse models are much less stable than non-sparse ones
\cite{varoquaux2012icml}.

\paragraph{Prediction power -- stability tradeoff}
The choice of decoder with the best predictive performance might not give
the most stable weight maps, as seen by comparing figures
\ref{fig:parameter_tuning} and \ref{fig:stability}. Figure
\ref{fig:tradeoff} shows the prediction--stability tradeoff for
different decoders and different parameter-tuning strategies.
Overall, SVM and logistic-regression perform similarly and the dominant
effect is that of the regularization: non-sparse, $\ell_2$-penalized,
models are much more stable than sparse, $\ell_1$-penalized, models. For
non-sparse models, averaging stands out as giving a large gain in
stability albeit with a decrease of a couple of percent in
prediction accuracy compared to using
$C=1$ or $C=1000$, which gives good prediction and stability (figures
\ref{fig:parameter_tuning} and \ref{fig:tradeoff}). With sparse
models, averaging offers a
slight edge for stability and, for SVM performs also well in prediction.
$C=1000$ achieves low stability (figure
\ref{fig:stability}), low prediction power (figure
\ref{fig:parameter_tuning}), and a poor tradeoff.

\ref{sec:stability_prediction_trends_appendix} shows trends
on datasets where the prediction is easy or not. For non-sparse models,
averaging brings a larger gain in stability when prediction accuracy
is large. Conversely, for sparse models, it is more beneficial to average
in case of poor prediction accuracy.

Note that these experimental results are for common neuroimaging
settings, with variance-normalization and univariate feature screening.


\section{Discussion and conclusion: lessons learned}

Decoding seeks to establish a predictive link between
brain images and behavioral or phenotypical variables.
Prediction is intrinsically a notion related to new data, and therefore
it is hard to measure. Cross-validation is the tool of choice to assess
performance of a decoder and to tune its parameters.
The strength of cross-validation is that it relies on few
assumptions and probes directly the ability to predict, unlike other
model-selection procedures --\emph{eg} based on information theoretic
or Bayesian criteria. However, it is limited by the small sample sizes
typically available in neuroimaging\footnote{There is a trend in
  inter-subject analysis to acquire databases with a larger number of
  subjects, \emph{eg} ADNI, HCP. Conclusions of our empirical study
  might not readily transfer to these settings.}.

\paragraph{An imprecise assessment of prediction}
The imprecision on the estimation of decoder performance by
cross-validation is often underestimated. Empirical confidence
intervals of cross-validated accuracy measures
typically extend more than 10 points up and down (figure \ref{fig:cv_error}
and \ref{fig:cv_strategies}). Experiments on MRI (anatomical and
functional), MEG, and simulations consistently exhibit these large
error bars due to data scarcity.
Such limitations should be kept in mind for
many MVPA practices that use predictive power as a form of hypothesis
testing --\emph{eg} searchlight \cite{kriegeskorte2006} or testing for
generalization \cite{knops2009recruitment}-- and it is recommended to use
permutation to define the null hypothesis \cite{kriegeskorte2006}. In
addition, in the light of cross-validation variance, methods publications
should use several datasets to validate a new model.

\paragraph{Guidelines on cross-validation}
Leave-one-out cross-validation should be avoided, as it yields more variable results. Leaving out
blocks of correlated observations, rather than individual observations,
is crucial for non-biased estimates. Relying on repeated random splits
with 20\% of the data enables better estimates with less computation
by increasing the number
of cross-validations without shrinking the size of the test set.

\paragraph{Parameter tuning}


Selecting optimal parameters can improve prediction and change
drastically the aspects of weight maps
(Fig.\,\ref{fig:maps}).

However,
our empirical study shows that for variance-normalized neuroimaging data,
non-sparse decoders ($\ell_2$-penalized) are
only weakly sensitive to the choice of their parameter, particularly for the
SVM. As a result, relying on the default value of the parameter often
outperforms parameter tuning by nested cross-validation. Yet, such
parameter tuning tends to improve the stability of the maps.
For sparse decoders ($\ell_1$-penalized), default parameters also give
good prediction performance. However, parameter tuning
with model averaging increases stability and can lead to better prediction.
Note that it is often useful to variance normalize the data (see
\ref{sec:not_standardized}).

\paragraph{Concluding remarks}
Evaluating a decoder is hard. Cross-validation should not be considered as a
silver bullet. Neither should prediction performance be the only metric. To
assess decoding accuracy, best practice is to use
repeated learning-testing with 20\% of the data left out, while keeping
in mind the large variance of the procedure. Any
parameter tuning should be performed in nested cross-validation, to limit
optimistic biases. Given the variance that arises from small samples, the
choice of decoders and their parameters should be guided by
several datasets.

Our extensive empirical validation (31 decoding tasks, with 8 datasets and
almost 1\,000 validation splits with nested
cross-validation) shows that sparse models, in particular $\ell_1$
SVM with model averaging, give better prediction but worst
weight-maps stability than non-sparse classifiers. If stability of weight
maps is important, non-sparse SVM with $C=1$ appears to be a good choice.
Further work calls for empirical studies of decoder performance with
more datasets, to reveal factors of the dataset that could guide better
the choice of a decoder for a given task.

\subsection*{Acknowledgments}

This work was supported by the EU FP7/2007-2013 under grant
agreement no. 604102 (HBP). Computing resource were provided by
the NiConnect project (ANR-11-BINF-0004\_NiConnect) and an
Amazon Webservices Research Grant.
The authors would like to thank the developers of
nilearn\footnote{\url{https://github.com/nilearn/nilearn/graphs/contributors}}, 
scikit-learn\footnote{\url{https://github.com/scikit-learn/scikit-learn/graphs/contributors}}
and MNE-Python\footnote{\url{https://github.com/mne-tools/mne-python/graphs/contributors}}
for continuous efforts in producing high-quality tools crucial for this
work.

In addition, we acknowledge useful feedback from Russ Poldrack on
the manuscript.


\small
\bibliographystyle{model1b-num-names}
\bibliography{biblio}

\appendix
\renewcommand{\thefigure}{A\arabic{figure}}
\setcounter{figure}{0}
\renewcommand{\thetable}{A\arabic{table}}
\setcounter{table}{0}

\section{Experiments on simulated data}


\label{sec:simulated_data}

\subsection{Dataset simulation}

We generate data with samples from two classes, each described by a
Gaussian of identity covariance in 100 dimensions. The classes are
centered respectively on vectors $(\mu, \dots, \mu)$ and 
$(-\mu, \dots, -\mu)$ where $\mu$ is a parameter adjusted to control the
separability of the classes. With larger $\mu$
the expected predictive accuracy would be higher. In addition, to mimic
the time dependence in neuroimaging data we apply a Gaussian smoothing
filter in the sample direction on the noise ($\sigma = 2$).
Code to reproduce the simulations can be found on
\url{https://github.com/GaelVaroquaux/cross_val_experiments}.

\begin{figure}
\vbox{}\hfill
\includegraphics[width=.4\linewidth]{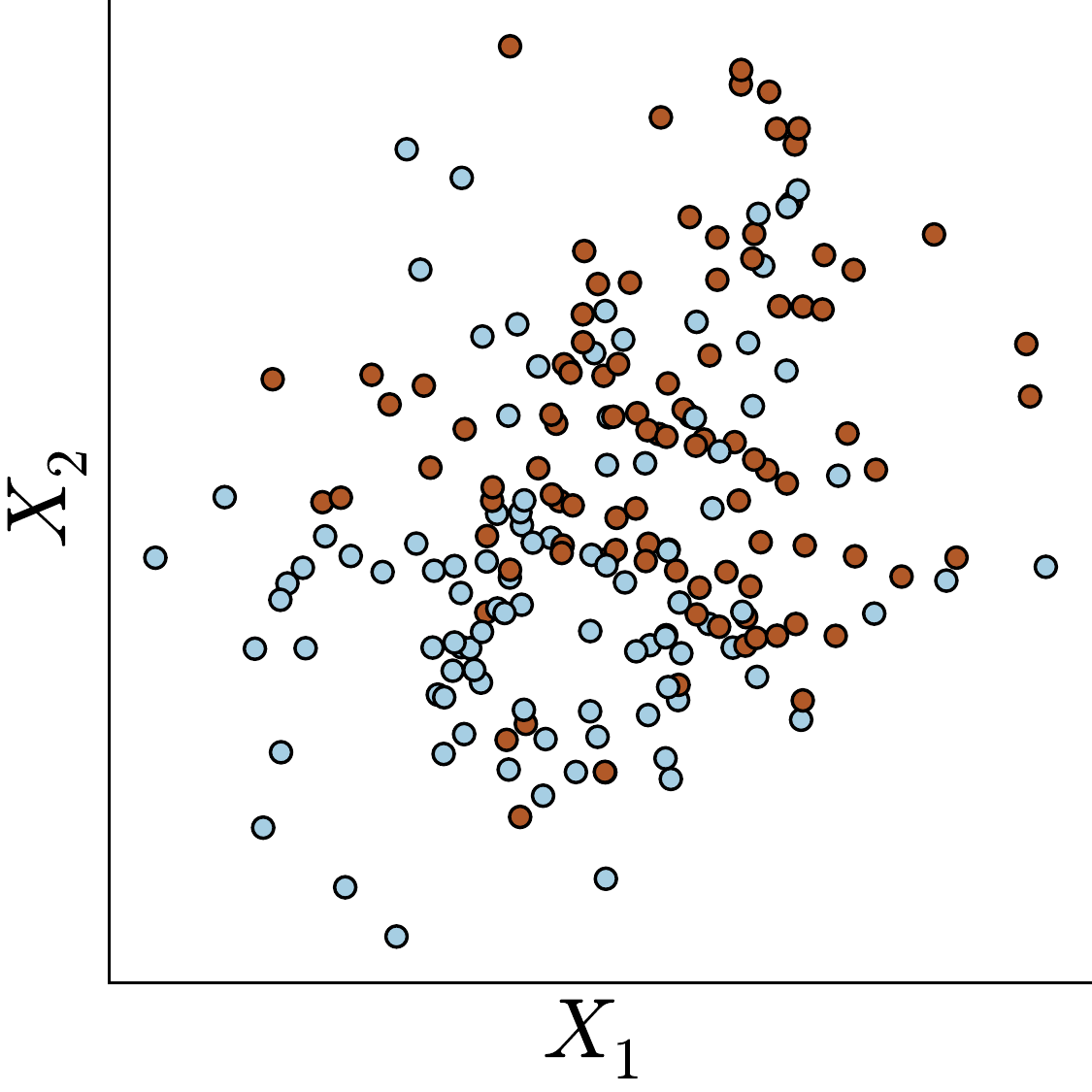}%
\hfill
\hfill
\hfill
\includegraphics[width=.4\linewidth]{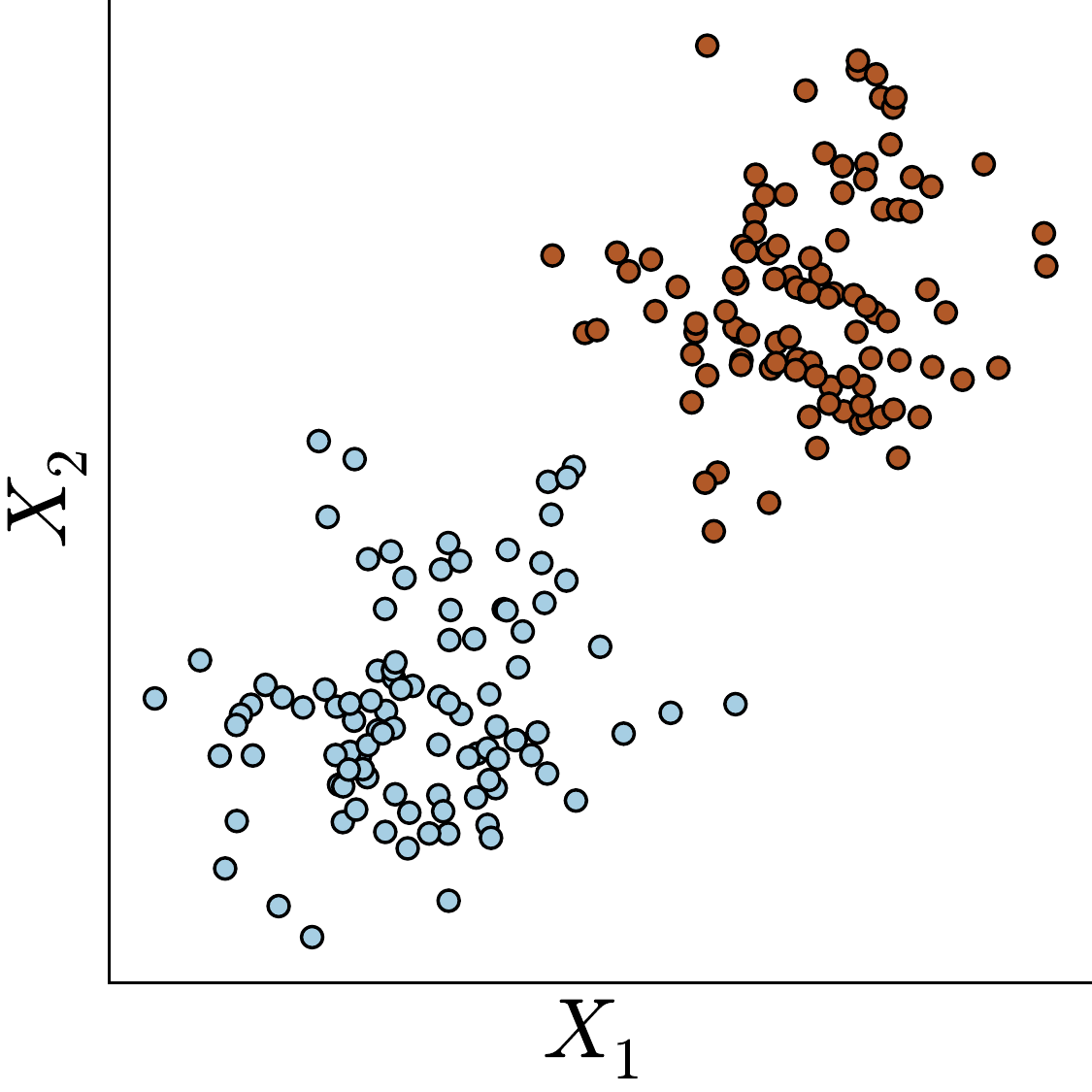}%
\hfill\vbox{}

{\footnotesize\bfseries\sffamily\qquad\qquad Low separability \hfill High separability
\quad\qquad\vbox{}}%
\vspace*{-1em}

\vbox{}\hfill
\includegraphics[width=.4\linewidth]{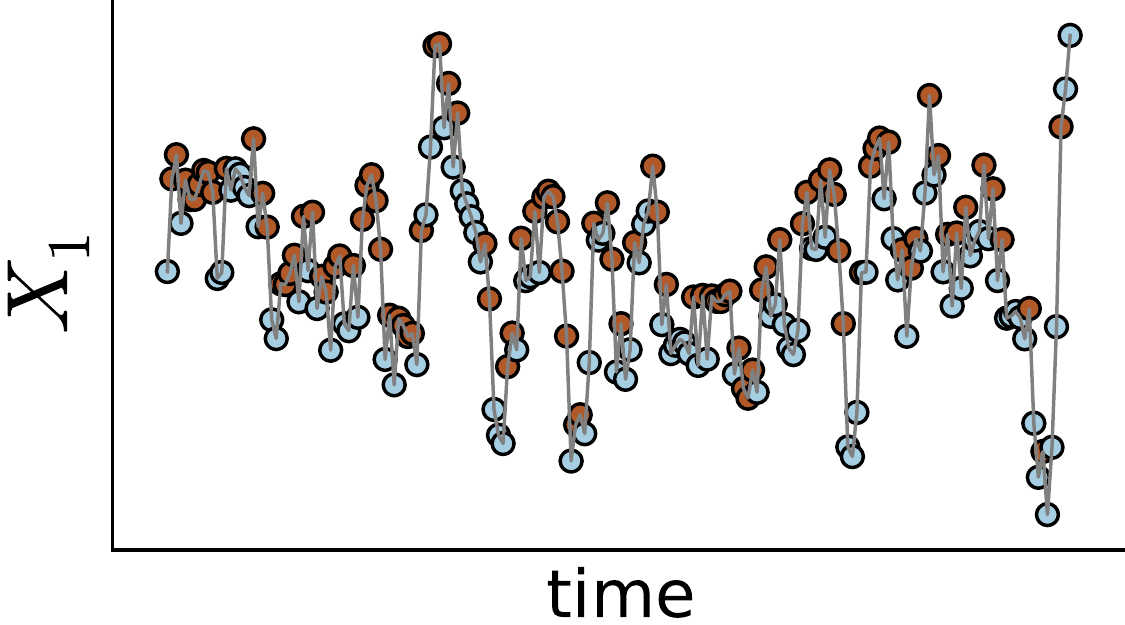}%
\hfill
\hfill
\hfill
\includegraphics[width=.4\linewidth]{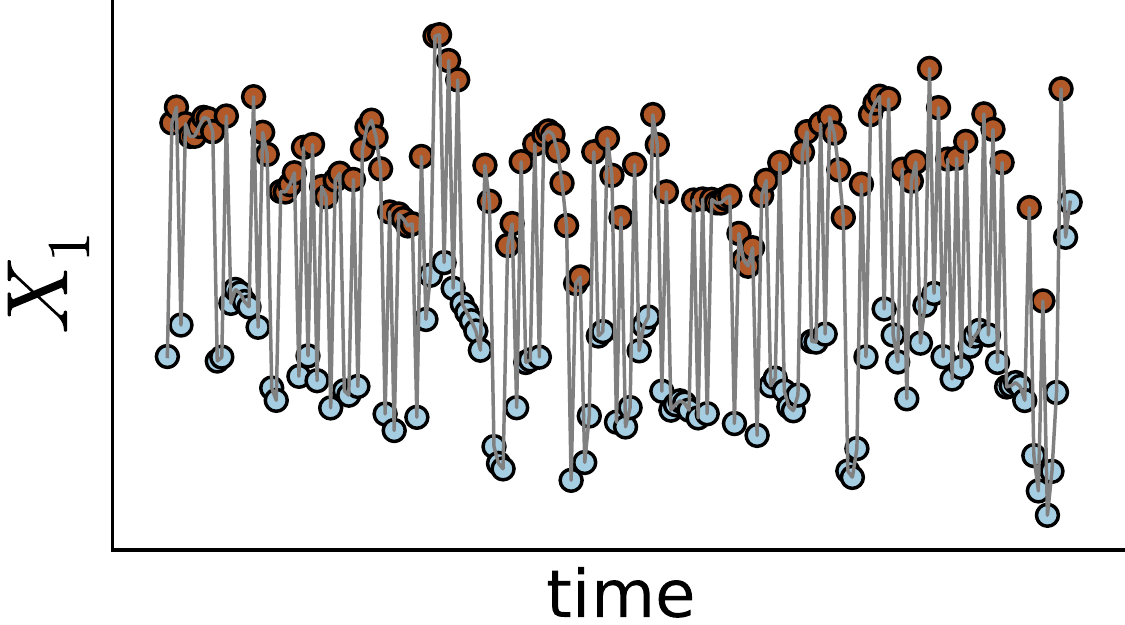}%
\hfill\vbox{}
\caption{\textbf{Simulated data for different levels of separability}
between the two classes (red and blue  circles). Here, to simplify
visualization, the data are generated in 2D (2 features), unlike the actual
experiments, which are performed on 100 features.
\textbf{Top}: The feature space.
\textbf{Bottom}: Time series of the first feature. Note that the noise is
correlated timewise, hence successive data points show similar shifts.
\label{fig:simulated_datasets}}
\end{figure}

We produce different datasets with predefined separability by
varying\footnote{the values we explore for $\mu$ were chosen empirically
to vary classification accuracy from 60\% to 90\%.} $\mu$ in $(.05,\;
.1,\; .2)$. Figure \ref{fig:simulated_datasets} shows two of these
configurations. 

\subsection{Experiments: error varying separability}
\label{sec:simdataclassifier}

Unlike with a brain imaging datasets, simulations open the door to
measuring the actual prediction performance of a classifier, and
therefore comparing it to the cross-validation measure.

For this purpose, we generate a pseudo-experimental data with 200 train
samples, and a separate very large test set, with 10\,000 samples. The
train samples correspond to the data available during a neuroimaging
experiment, and we perform cross-validation on these. We apply the
decoder on the test set. The large number of test samples provides a good
measure of prediction power of the decoder \cite{arlot2010}. We use the
same decoders as for brain-imaging data and repeat the whole procedure
100 times. For cross-validation strategies that rely on sample blocks
--as with sessions--, we divide the data in 10 continuous blocks.

\begin{figure}\smallskip
{\small\sffamily
Cross-validation\hfill {\bfseries Accuracy measured~by cross-validation}\hfill\vbox{}\vspace*{-.4ex}

strategy\hfill\hfill {\bfseries  for different separability~~}\hfill\vbox{}}
\includegraphics[width=\linewidth]{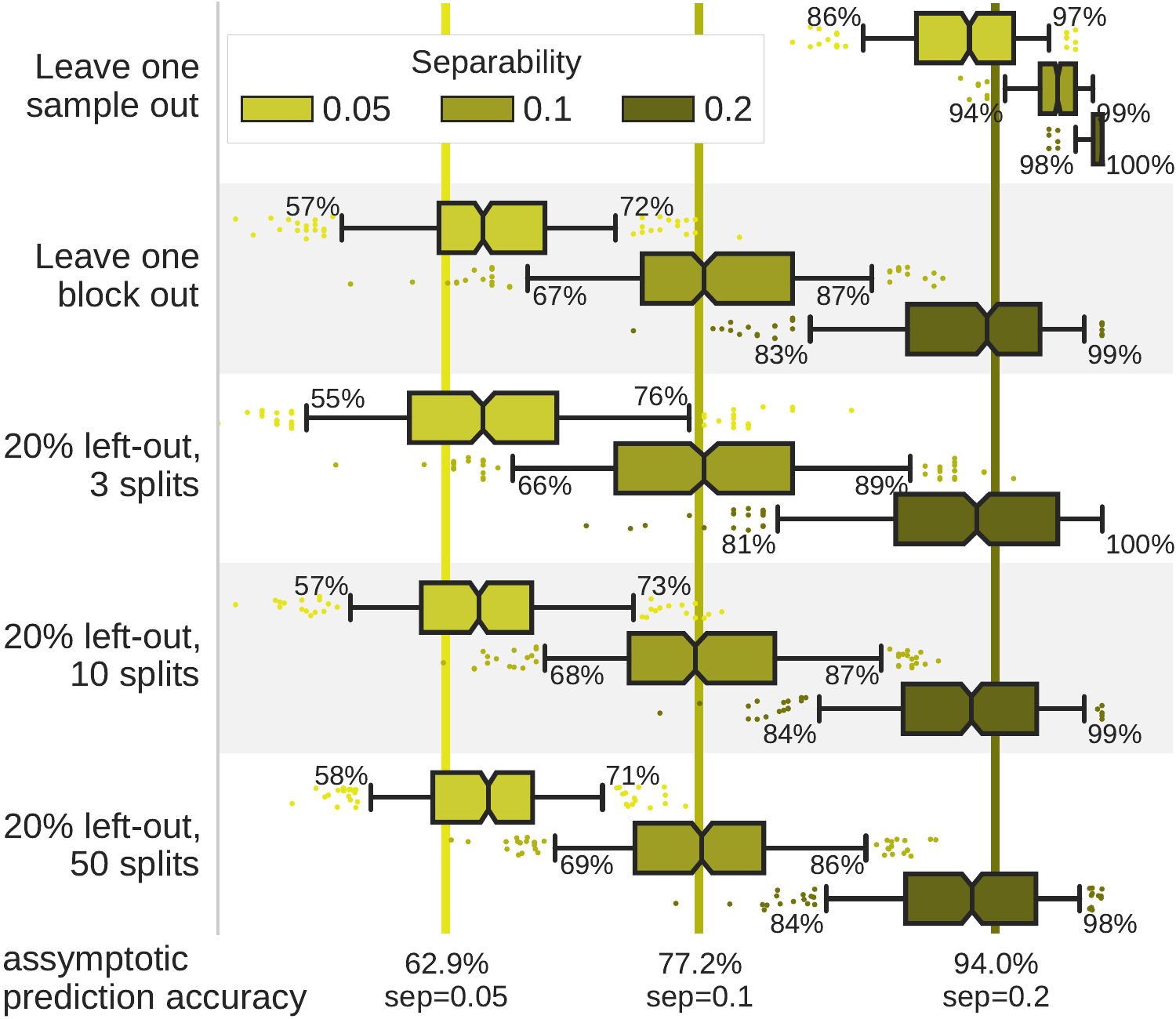}

\caption{\textbf{Cross-validation measures on simulations}.
    Prediction accuracy, as measured by cross-validation (box plots)
    and on a very large test set (vertical lines) for different
    separability on the simulated data and for different
    cross-validation strategies: leave one sample out,
    leave one block of samples out (where the block is the natural
    unit of the experiment: subject or session), or
    random splits leaving out 20\% of the blocks as test data, with 3, 10, or
    50 random splits.
    The box gives the quartiles, while the whiskers give
    the 5 and 95 percentiles.
    Note that here leave-one-block-out is similar of 10 splits of 10\% of
    the data.
    \label{fig:cv_simulations}}
\end{figure}

\paragraph{Results}
Figure \ref{fig:cv_simulations} summarizes the cross-validation measures
for different values of separability. Beyond the effect of the
cross-validation strategy observed on other figures, the effect of the
separability, \emph{ie} the true prediction accuracy is also visible.
Setting aside the leave-one-sample-out cross-validation strategy, which
is strongly biased by the correlations across the samples, we see that
all strategies tend to be biased positively for low accuracy and
negatively for high accuracy. This observation is in accordance with
trends observed on figure \ref{fig:cv_error}.

\section{Comparing parameter-tuning strategies}

\begin{figure}[t]
\setlength{\mylength}{18cm}%
\hspace*{-.1ex}%
\includegraphics[height=.238\mylength,
		 trim={0 .4cm 0 0}]{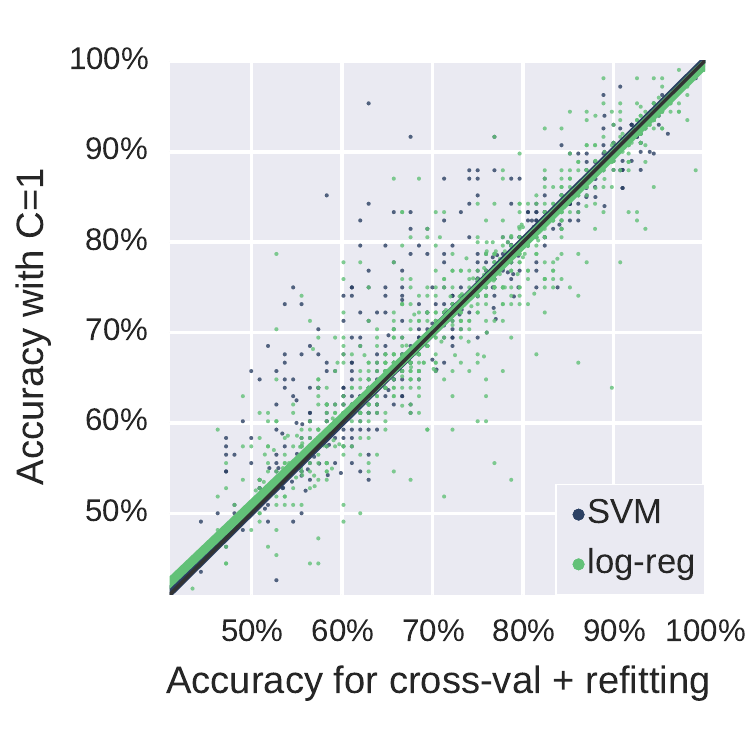}%
\hspace*{-.3ex}%
\includegraphics[height=.238\mylength,
		 trim={0 .4cm 0 0}]{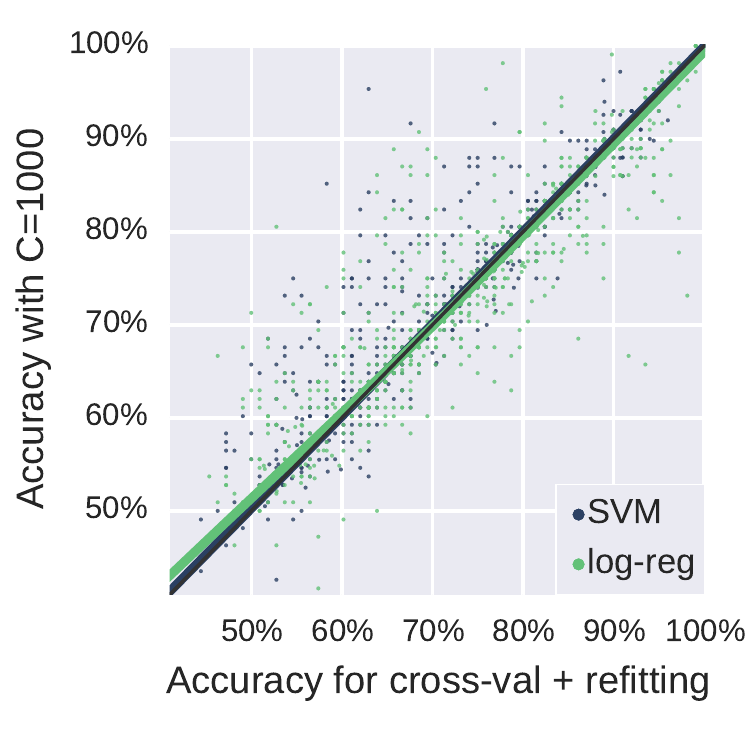}%
\llap{\raisebox{.22\mylength}{%
    \rlap{\small\bfseries\sffamily Non-sparse models}}%
    \hspace*{.9\linewidth}%
    }%
\hspace*{-.1ex}%

\hspace*{-.1ex}%
\includegraphics[height=.238\mylength,
		 trim={0 .4cm 0 0}]{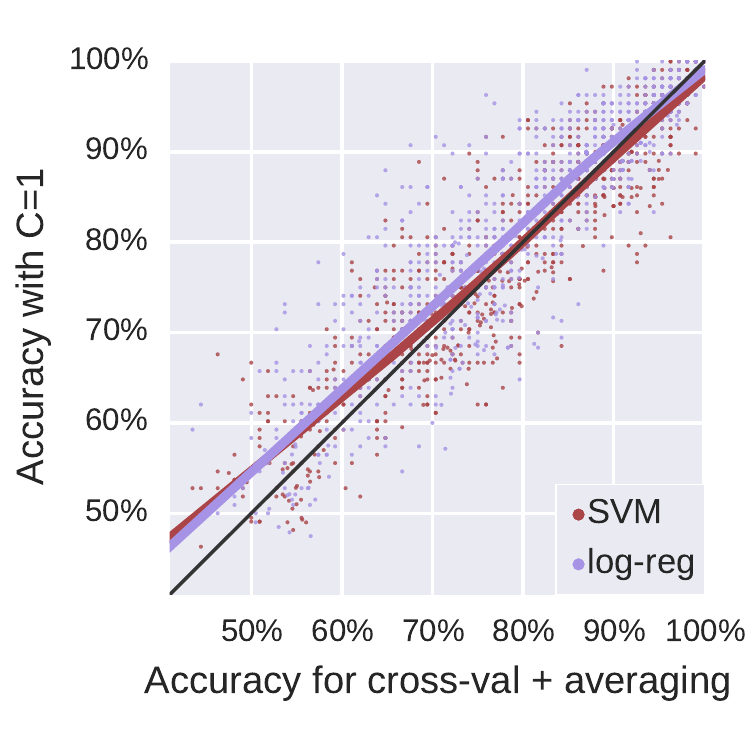}%
\hspace*{-.3ex}%
\includegraphics[height=.238\mylength,
		 trim={0 .4cm 0 0}]{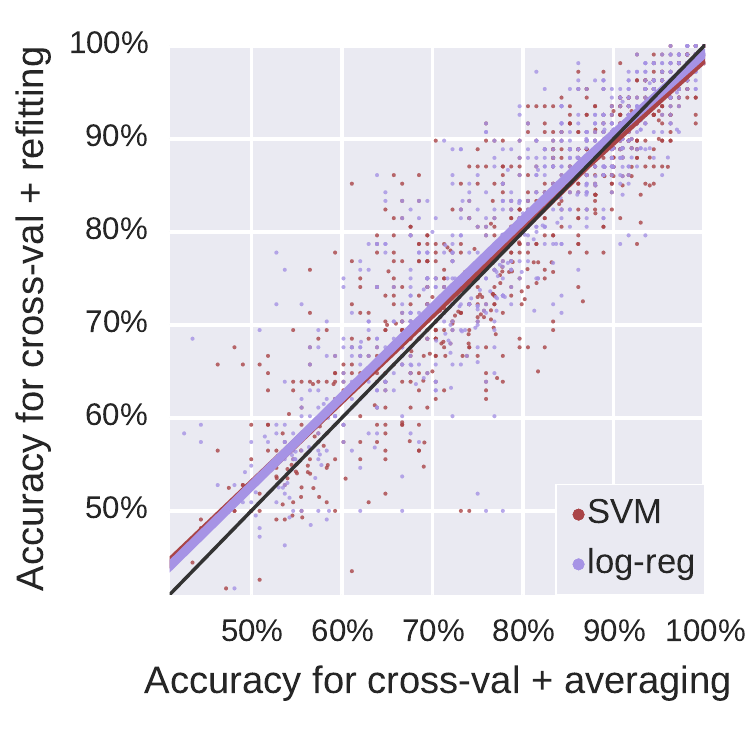}%
\llap{\raisebox{.218\mylength}{%
    \rlap{\small\bfseries\sffamily Sparse models}}%
    \hspace*{.9\linewidth}%
    }µ
\hspace*{-.1ex}%

\caption{\textbf{Comparing parameter-tuning strategies on prediction
accuracy}. Relating pairs
of tuning strategies. Each dot corresponds to a given dataset, task, and
validation split.
The line is an indication of the tendency, using
a lowess non-parametric local regression.
The top row summarizes results for non-sparse models, SVM $\ell_2$ and
logistic regression $\ell_2$; while the bottom row gives results for
sparse models, SVM $\ell_1$ and logistic regression $\ell_1$.
\label{fig:parameter_tuning_ext}}
\end{figure}

Figure \ref{fig:parameter_tuning_ext} shows pairwise comparisons of
parameter-tuning strategies, in sparse and non-sparse situations, for the
best-performing options. In particular, it investigates when different
strategies should be preferred. The trends are small. Yet, it appears
that for low predictive power, setting C=1 in non-sparse models is
preferable to cross-validation while for high predictive power,
cross-validation is as efficient. This is consistent with results in
figure \ref{fig:cv_error} showing that cross-validation is more
reliable to measure prediction error in situations with a good accuracy
than in situations with a poor accuracy. Similar trends can be found when
comparing to C=1000. For sparse models, model averaging can be preferable
to refitting. We however find that for low prediction accuracy it is
favorable to use C=1, in particular for logistic regression.

Note these figures show points related to different studies and
classification tasks. The trends observed are fairly homogeneous and
there are not regions of the diagram that stand out. Hence, the
various conclusions on the comparison of decoding strategies are driven
by all studies.

\section{Results without variance-normalization}

\label{sec:not_standardized}

\begin{figure}[!t]

\centerline{%
\includegraphics[height=.65\linewidth,
		 trim={0 .4cm 0 0}]{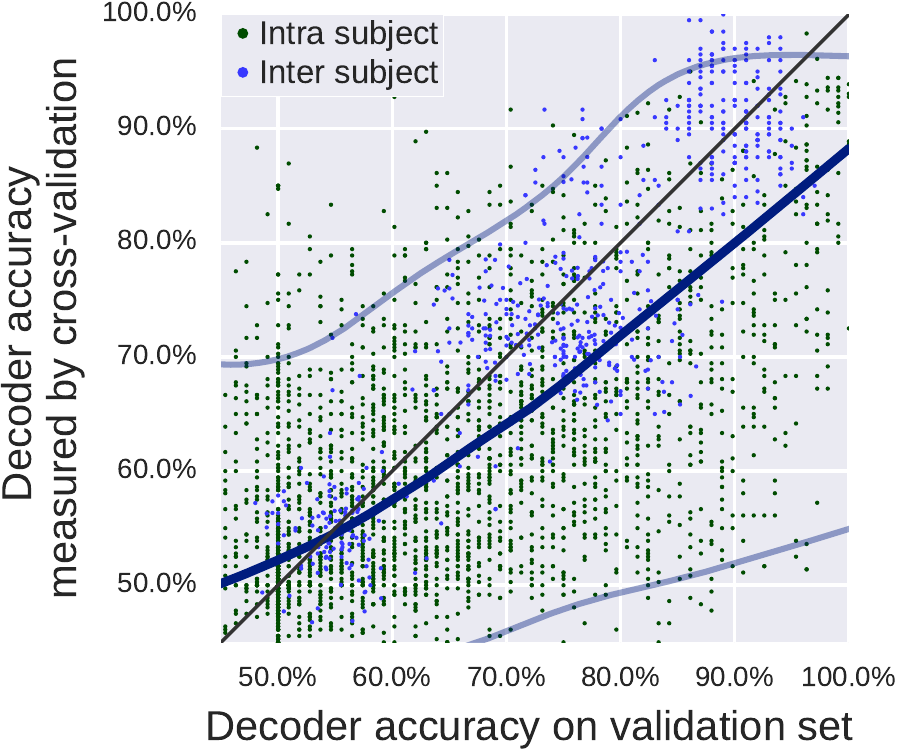}%
}%

\caption{\textbf{Prediction error: cross-validated versus validation
set}. Each point is a measure of predictor error measure in the inner
cross-validation loop, or in the left-out validation dataset
for a model refit using the best parameters.
The line is an indication of the tendency, using a lowess non-parametric
local regression.
\label{fig:cv_error_raw}}
\end{figure}

\begin{figure}[!t]

\setlength{\mylength}{20cm}%
\includegraphics[height=.238\mylength,
		 trim={0 .4cm 0 0}]{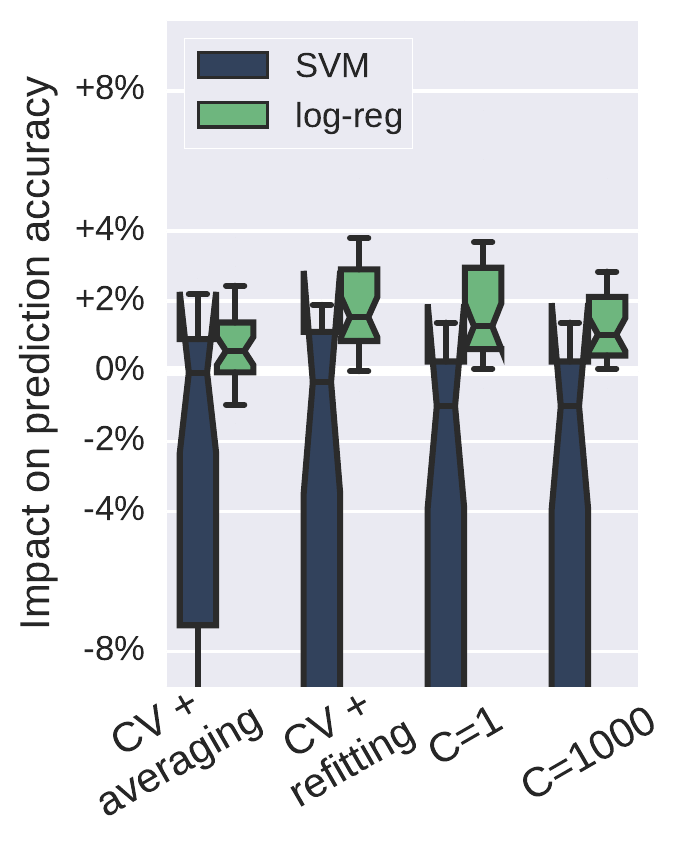}%
\llap{\raisebox{.234\mylength}{%
    \rlap{\small\bfseries\sffamily Non-sparse models}}%
    \hspace*{.34\linewidth}%
    }%
\hfill%
\includegraphics[height=.238\mylength,
		 trim={0 .4cm 0 0}]{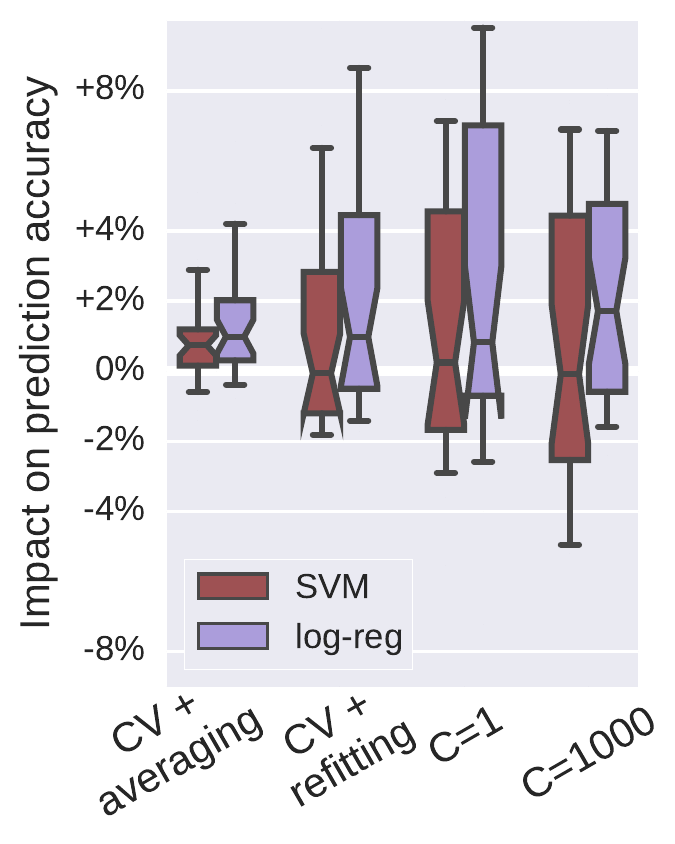}%
\llap{\raisebox{.234\mylength}{%
    \rlap{\small\bfseries\sffamily Sparse models}}%
    \hspace*{.34\linewidth}%
    }

\caption{\textbf{Impact of the parameter-tuning strategy on the
prediction accuracy without feature standardization}: for each strategy, difference to the mean stability of
the model weights across validation splits.
\label{fig:tuning_parameters_raw}}

\hfill
\bigskip
\bigskip
\medskip

\setlength{\mylength}{20cm}%
\includegraphics[height=.238\mylength,
		 trim={0 .4cm 0 0}]{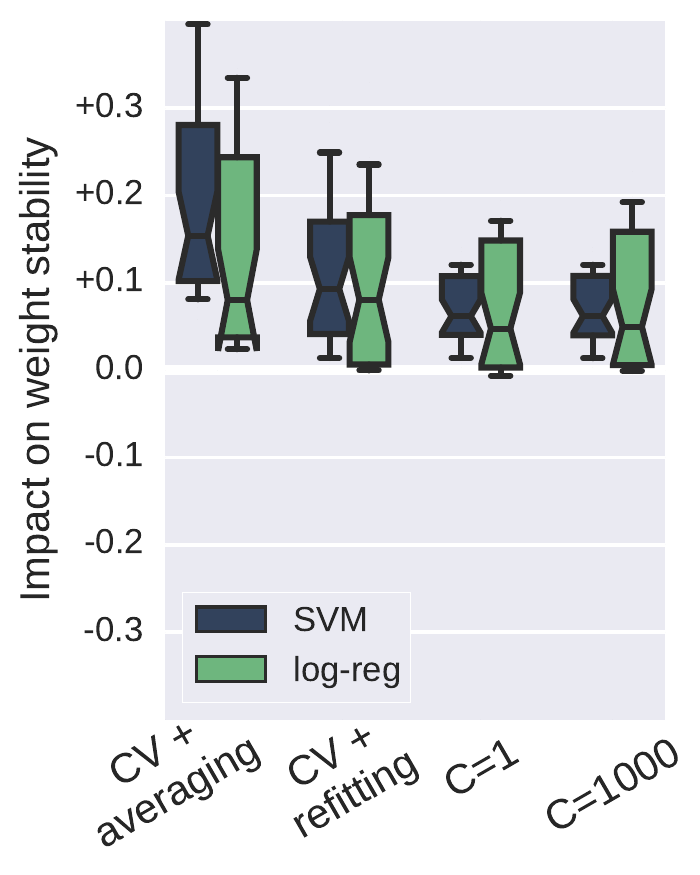}%
\llap{\raisebox{.234\mylength}{%
    \rlap{\small\bfseries\sffamily Non-sparse models}}%
    \hspace*{.34\linewidth}%
    }%
\hfill%
\includegraphics[height=.238\mylength,
		 trim={0 .4cm 0 0}]{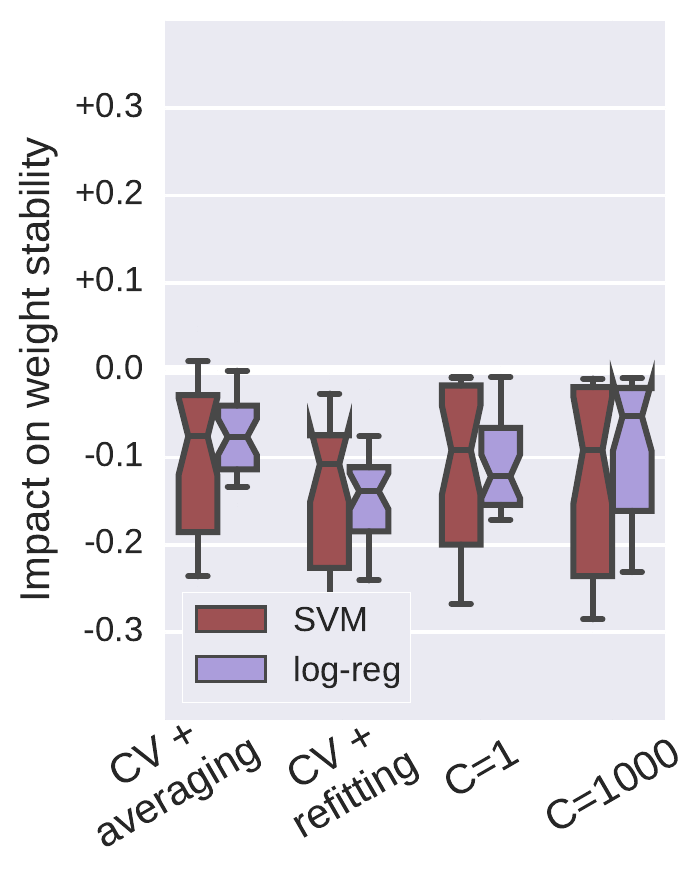}%
\llap{\raisebox{.234\mylength}{%
    \rlap{\small\bfseries\sffamily Sparse models}}%
    \hspace*{.34\linewidth}%
    }

\caption{\textbf{Impact of the parameter-tuning strategy on stability of
weights without feature standardization}: for each strategy, difference to the mean stability of
the model weights across validation splits.
\label{fig:stability_raw}}
\end{figure}

Results without variance normalization of the voxels are given in figure
\ref{fig:cv_error_raw} for the correspondence between error
measured in the inner cross-validation loop, figure
\ref{fig:tuning_parameters_raw} for the effect of the choice of a
parameter-tuning strategy on the prediction performance,
and figure \ref{fig:stability_raw} for the effect on the stability of the
weights.

Cross-validation on non variance-normalized neuroimaging data is not more
reliable than on variance-normalized data (figure
\ref{fig:cv_error_raw}). However, parameter tuning by nested
cross-validation is more important than on variance-normalized data for
good prediction (figure \ref{fig:tuning_parameters_raw}). This difference
can be explained by the fact that variance normalizing makes dataset more
comparable to each others, and thus a default value of parameters is more
likely to work well.

In conclusion, variance normalizing the data can be important, in
particular with non-sparse SVM.

\section{Stability--prediction trends}

\label{sec:stability_prediction_appendix}

\subsection{Details on stability--prediction results}

\begin{figure*}
\begin{minipage}{.235\linewidth}
\caption{\textbf{Prediction -- stability tradeoff}
This figures gives the data points behind \ref{fig:tradeoff}, reporting
the stability of the weights, relative to the split's average,
as a function of the delta in prediction accuracy.
Each point is corresponds to a specific prediction task in our study.
\label{fig:tradeoff_details}}
\end{minipage}
\hfill%
\begin{minipage}{.66\linewidth}
\setlength{\mylength}{.5\linewidth}%
\hspace*{-2ex}%
\includegraphics[height=\mylength,
		 trim={0 .42cm .1cm 0}]{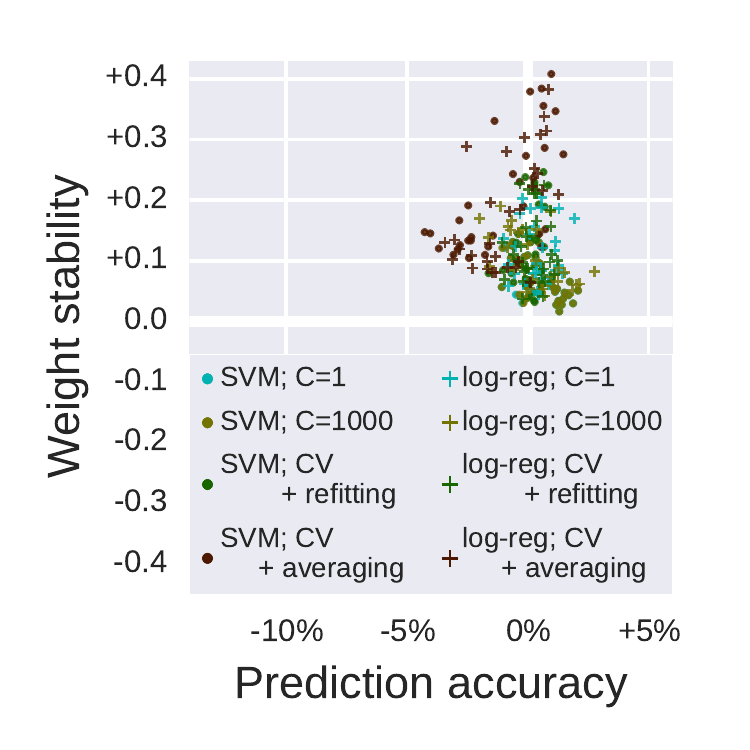}%
\llap{\raisebox{.93\mylength}{%
    \rlap{\small\sffamily Non-sparse models}}%
    \hspace*{.78\mylength}%
    }
\hspace*{-3ex}%
\hfill%
\includegraphics[height=\mylength,
		 trim={0 .42cm .1cm 0}]{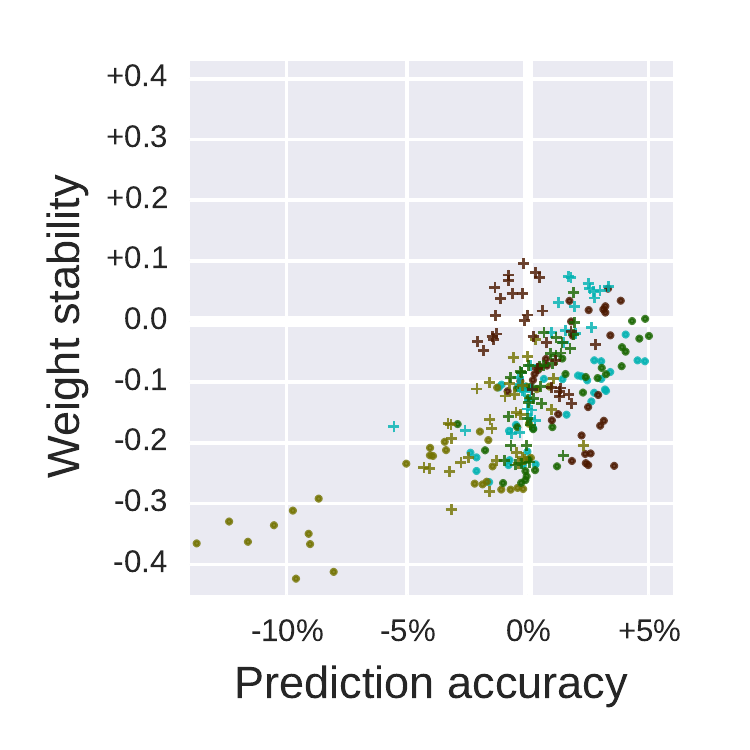}%
\llap{\raisebox{.93\mylength}{%
    \rlap{\small\sffamily Sparse models}
    \hspace*{.75\mylength}%
}%
    }%
\hspace*{-2.5ex}
\end{minipage}%
\hfill%
\vbox{}%
\end{figure*}

Figure \ref{fig:tradeoff} summarizes the effects of the decoding strategy
on the prediction -- stability tradeoff. On figure 
\ref{fig:tradeoff_details}, we give the data points that underly this
summary.

\subsection{Prediction and stability interactions}

\label{sec:stability_prediction_trends_appendix}

\begin{figure*}
\begin{minipage}{.235\linewidth}\vspace*{-.5ex}
\caption{\textbf{Prediction -- stability tradeoff}
The figure reports for each dataset and task the stability of the weights
as a function of the prediction accuracy measured on the validation set,
with the different choices of decoders and parameter-tuning strategy.
The line is an indication of the tendency, using a lowess
local regression. The stability of the weights is the correlation across
validation splits.
\label{fig:tradeoff_trends}}
\end{minipage}%
\hfill%
\begin{minipage}{.75\linewidth}%
\setlength{\mylength}{18.2cm}%
\includegraphics[height=.238\mylength,
		 trim={0 .4cm 0 0}]{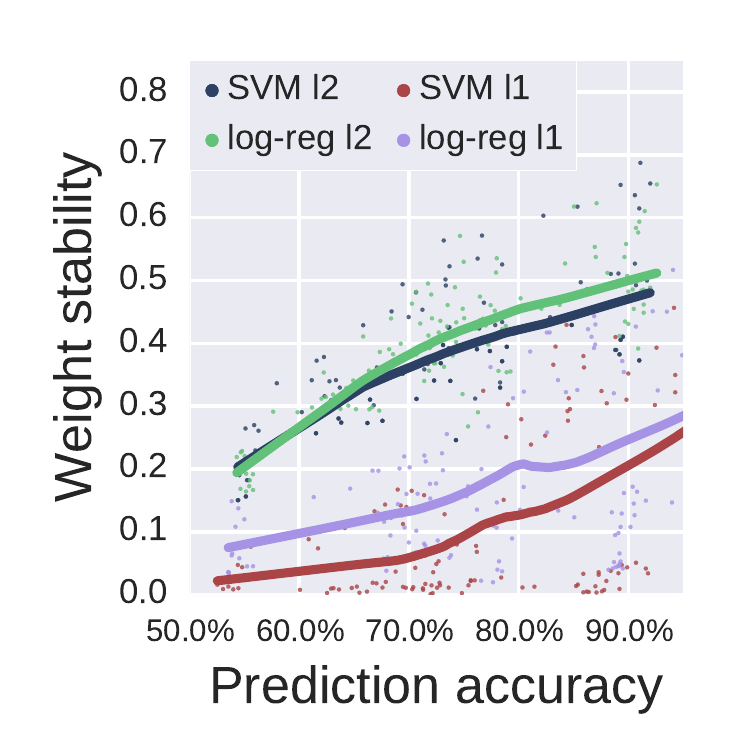}%
\llap{\raisebox{.22\mylength}{%
    \rlap{\small\sffamily All models}}%
    \hspace*{.25\linewidth}%
    }
\hspace*{-.5ex}\hfill%
\includegraphics[height=.238\mylength,
		 trim={0 .4cm 0 0}]{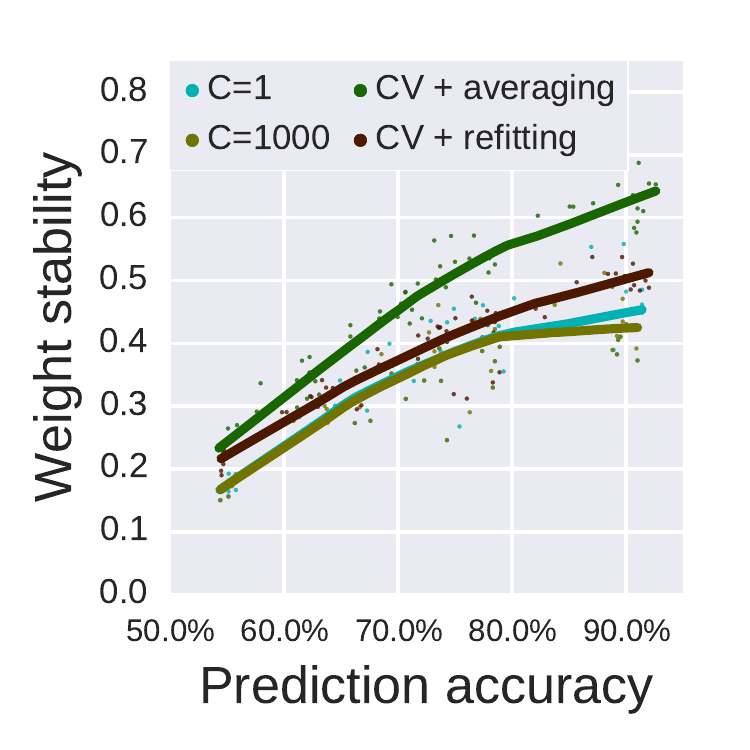}%
\llap{\raisebox{.22\mylength}{%
    \rlap{\small\sffamily Non-sparse models}}%
    \hspace*{.256\linewidth}%
    }
\hspace*{-2.5ex}%
\includegraphics[height=.238\mylength,
		 trim={0 .4cm 0 0}]{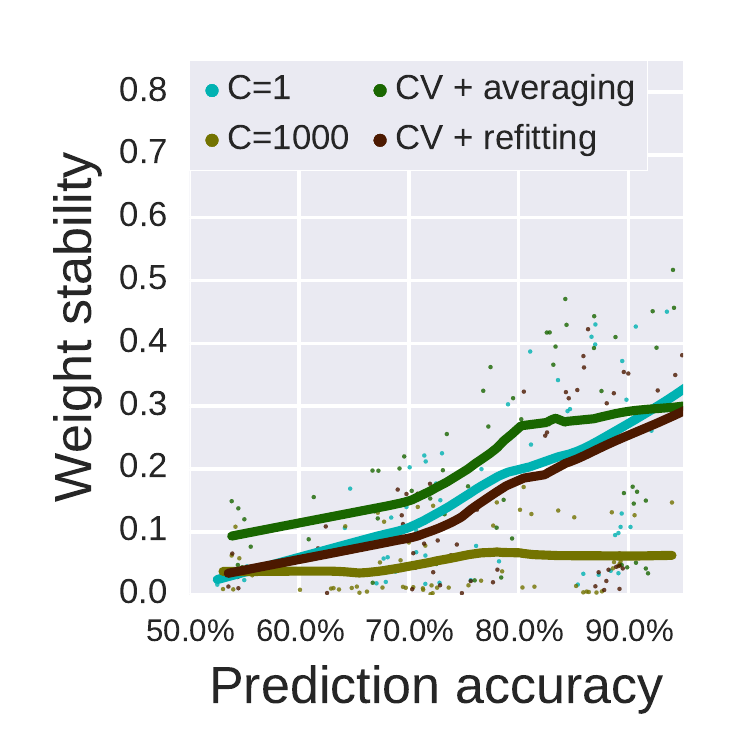}%
\llap{\raisebox{.22\mylength}{%
    \rlap{\small\sffamily Sparse models}
    \hspace*{.24\linewidth}%
}%
    }%
\end{minipage}
\end{figure*}

Figure \ref{fig:tradeoff} captures the effects of the decoding strategy
across all datasets. However, some classification tasks are easier or
more stable than others.

We give an additional figure, figure \ref{fig:tradeoff_trends}, showing
this interaction between classification performance and the best
decoding strategy in terms of weight stability. The main factor of
variation of the prediction accuracy is the choice of dataset, \emph{ie}
the difficulty of the prediction task.

Here again, we see that the most important choice is that of the penalty:
logistic regression and SVM have overall the same behavior. In terms
of stability of the weights, higher prediction accuracy does correspond
to more stability, except for overly-penalized sparse model (C=1000).
For non-sparse models, model averaging after cross-validation is
particularly beneficial in good prediction situations.

\section{Details on datasets used}

\label{sec:table}%

\begin{table*}%
\centering%
\definecolor{Gray}{gray}{0.92}%
\definecolor{LightCyan}{rgb}{0.88,1,1}%
\def\arraystretch{1.2}%
\footnotesize%
%
\begin{tabular}{lccc|r|cc}
& & & \multirow{2}{*}{\parbox{1.6cm}{\#\,blocks (sess./subj.)}} &  & \multicolumn{2}{c}{mean accuracy} \\
Dataset & Description & \# samples & & Task                  & SVM $\ell_2$ & SVM $\ell_1$   \\
\hline
\rowcolor{Gray} &       &&       & bottle\,/\,scramble   & $75\%$  & $86\%$  \\
\rowcolor{Gray} &       &&       & cat\,/\,bottle        & $62\%$  & $69\%$  \\
\rowcolor{Gray} &       &&       & cat\,/\,chair         & $69\%$  & $80\%$  \\
\rowcolor{Gray} &       &&       & cat\,/\,face          & $65\%$  & $72\%$  \\
\rowcolor{Gray} &       &&       & cat\,/\,house         & $86\%$  & $95\%$  \\
\rowcolor{Gray} &       &&       & cat\,/\,scramble      & $83\%$  & $92\%$  \\
\rowcolor{Gray} &       &&       & chair\,/\,scramble    & $77\%$  & $91\%$  \\
\rowcolor{Gray} Haxby \cite{haxby2001}   
                & fMRI  &  209  & 12 sess.& chair\,/\,shoe        & $63\%$  & $70\%$  \\
\rowcolor{Gray} & 5 different subjects,\qquad\vbox{}
                        &&       & face\,/\,house        & $88\%$  & $96\%$  \\
\rowcolor{Gray} & leading to 5 experiments
	                &&       & face\,/\,scissors     & $72\%$  & $83\%$  \\
\rowcolor{Gray} &  per task\quad\qquad\qquad\qquad\vbox{}
		        &&       & scissors\,/\,scramble & $73\%$  & $87\%$  \\
\rowcolor{Gray} &       &&       & scissors\,/\,shoe     & $60\%$  & $64\%$  \\
\rowcolor{Gray} &       &&       & shoe\,/\,bottle       & $62\%$  & $69\%$  \\
\rowcolor{Gray} &       &&       & shoe\,/\,cat          & $72\%$  & $85\%$  \\
\rowcolor{Gray} &       &&       & shoe\,/\,scramble     & $78\%$  & $88\%$  \\
                &       &&       & consonant\,/\,scramble & $92\%$  & $88\%$  \\
\multirow{4}{*}{Duncan \cite{duncan2009consistency}}
& \multirow{2}{*}{fMRI,}   
& \multirow{4}{*}{196} & \multirow{4}{*}{49 subj.} & consonant\,/\,word             & $92\%$  & $89\%$  \\
& \multirow{2}{*}{across subjects} 
                       &&        & objects\,/\,consonant & $90\%$  & $88\%$  \\
                &      &&        & objects\,/\,scramble  & $91\%$  & $88\%$  \\
                &      &&        & objects\,/\,words     & $74\%$  & $71\%$  \\
                &      &&        & words\,/\,scramble    & $91\%$  & $89\%$  \\
\rowcolor{Gray}& fMRI &&& negative cue\,/\,neutral cue   & $55\%$  & $55\%$  \\
\rowcolor{Gray}Wager \cite{wager2008neural} 
& across subjects   & 390& 34 subj.  & negative rating\,/\,neutral rating
							& $54\%$  & $53\%$  \\
\rowcolor{Gray}&    &&       & negative stim\,/\,neutral stim
							& $77\%$  & $73\%$  \\
Cohen (ds009) & fMRI& 80 & 24 subj.  & successful\,/\,unsuccessful stop
							& $67\%$  & $63\%$  \\
& across subjects   &       &&                                   &    & \\
\rowcolor{Gray} Moran \cite{moran2012social}  
             & fMRI    & 138 & 36 subj. & false picture\,/\,false belief    & $72\%$   & $71\%$ \\
\rowcolor{Gray}& across subjects &&                     &         & & \\
\multirow{3}{*}{Henson \cite{henson2002face}}  
& \multirow{2}{*}{fMRI}     
& \multirow{3}{*}{286}       &\multirow{3}{*}{16 subj.}  & famous \,/\, scramble      & $77\%$  & $74\%$ \\
& \multirow{2}{*}{across subjects}  &&  & famous \,/\, unfamiliar
							  & $54\%$  & $55\%$ \\
&                                   &&  & scramble \,/\, unfamiliar
							  & $73\%$  & $70\%$ \\
\rowcolor{Gray} Knops \cite{knops2009recruitment}  
& fMRI,     & 114 & 19 subj. & right field \,/\, left field     & $79\%$  & $73\%$   \\
\rowcolor{Gray} 
& across subjects   &  &&  &   &  \\ 
Oasis \cite{marcus2007}  
& VBM       & 403 & 52 subj.& Gender discrimination   &  $77\%$  & $75\%$  \\
\rowcolor{Gray} &                              & 299 & 52 subj.& faces \,/\, tools &  $81\%$  & $78\%$  \\ 
\rowcolor{Gray} HCP \cite{larson2013adding}& MEG working memory & 223 & 52 subj.& target\,/\, non-target &  $58\%$  & $72\%$  \\
\rowcolor{Gray} &  across trials & 135 & 52 subj.& target \,/\, distractor &  $54\%$  & $53\%$  \\ 
\rowcolor{Gray} &                & 239 & 52 subj.& distractor \,/\, non-target &  $55\%$  & $67\%$  \\ 

\end{tabular}

\caption{\textbf{The different datasets and tasks}\label{tab:results}.
We report the prediction performance on the validation test for
parameter tuning by 10 random splits followed by refitting using the best
parameter.
}
\end{table*}

Table \ref{tab:results} lists all the studies used in our experiments, as
well as the specific prediction tasks. In the Haxby dataset
\cite{haxby2001} we use various pairs of visual stimuli, with differing
difficulty. We excluded pairs for which decoding was unsuccessful, such
as scissors versus bottle).

\section{Details on preprocessing}

\subsection{fMRI data}

\label{sec:fmri_data_analysis}

\paragraph{Intra-subject prediction}
For intra-subject prediction, we use the Haxby dataset \cite{haxby2001}
as provided from the PyMVPA \cite{hanke2009pymvpa} website
--\url{http://dev.pymvpa.org/datadb/haxby2001.html}. Details of the
preprocessing are not given in the original paper, beyond the fact that
no spatial smoothing was performed. We have not performed additional
preprocessing on top of this publicly-available dataset, aside from spatial
smoothing with an isotropic Gaussian kernel, FWHM of 6\,mm (nilearn 0.2,
Python 2.7).

\paragraph{Inter-subject prediction}
For inter-subject prediction, we use different datasets available on
openfMRI \cite{poldrack2013openfmri}. For all the datasets, we performed
standard preprocessing with SPM8\footnote{Wellcome Department of
Cognitive Neurology, \url{http://www.fil.ion.ucl.ac.uk/spm}}: in the
following order, slice-time correction, motion correction (realign),
corregistration of mean EPI on subject's T1 image, and normalization to
template space with unified segmentation on the T1 image. 
The preprocessing pipeline was orchestrated through the Nipype
processing infrastructure \cite{gorgolewski2011}.
For each subject, we then performed session-level GLM with a design
according to the individual studies, as described in the openfMRI
files, using Nipy (version 0.3, Python version 2.7)
\cite{millman2007analysis}.

\subsection{Structural MR data}

\label{sec:vbm_data_analysis}

For prediction from structural MR data, we perform Voxel Based
Morphometry (VBM) on the Oasis dataset \cite{marcus2007}. We use SPM8
with the following steps: segmentation of the white matter/grey
matter/CSF compartiments and estimation of the deformation fields with
DARTEL \cite{ashburner2011diffeomorphic}. The inputs for predictive models
is the modulated grey-matter intensity. The corresponding maps can be
downloaded with the dataset-downloading facilities of the nilearn
software (function {\tt nilearn.datasets.fetch\_oasis\_vbm}).

\subsection{MEG data}

\label{sec:meg_data_analysis}

The magneteoencephalography (MEG) data is from an N-back working-memory
experiment made available by the Human Connectome
Project~\cite{larson2013adding}. 
Data from 52 subjects and two runs was
analyzed using a temporal window approach in which all magnetic fields
sampled by the sensor array in a fixed time interval yield one variable
set (see for example ~\cite{sitt2014large} for event related potentials
in electroencephalography). Here, each of the two runs served as
validation set for the other run. For consistency, two-class decoding
problems were considered, focussing on either the image content (faces VS
tools) or the functional role in the working memory task (target VS
low-level and high-level distractors). This yielded in total four
classification analyses per subject. For each trial, the time window was
then constrained to 50 millisecond before and 300 millisecond after event
onset, emphasizing visual components.

All analyses were based on the
cleaned single-trial outputs obtained from the HCP ``tmegpreproc''
pipeline which provides cleaned segmented sensor space data. 
The MEG data that were recorded with a wholehead MAGNES 3600 (4D Neuroimaging,
San Diego, CA) magnetometer system in a magnetically shielded room.
Contamination by environmental magnetic fields
was accounted for by computing the residual MEG signal from concomitant
recordings of reference gradiometers and magnetometers located remotely
from the main sensor array. Data were
bandpass filtered between 1.3 and 150Hz using zero-phase forward and
everse Butterworth filters. Notch filters were then applied at
(59-61/119-121 Hz) to attenuate line noise artefacts.
Data segments contaminated by remaining environmental or system
artifacts were detected using a semi-automatic
HCP pipeline that takes into account the local and global variation as well as the
correlation structure of the data.
Independent component analysis based on the FastICA algorithm was then used
to estimate and suppress spatial patterns of cardiac and ocular artifacts.
Artifact related components were identified in a semi-automatic fashion assisted by 
comparisons with concomitantly recorded electrocardiogram (ECG) and electrooculogram (EOG).
These components were then projected out from the data.
Depending on the classification of bad channels performed by the HCP pipelines, the data contained fewer
than 248 sensors. For details on the HCP pipelines see~\citet{larson2013adding}
and the HCP reference manual.
The MEG data were accessed through the MNE-Python
software~\cite{gramfort2013meg, gramfort2014mne} and the MNE-HCP
library~\href{https://github.com/mne-tools/mne-hcp}.

\section{Performance on each classification task}

\begin{figure}[!b]%
\begin{minipage}{.89\paperwidth}%
    \includegraphics[width=\linewidth]{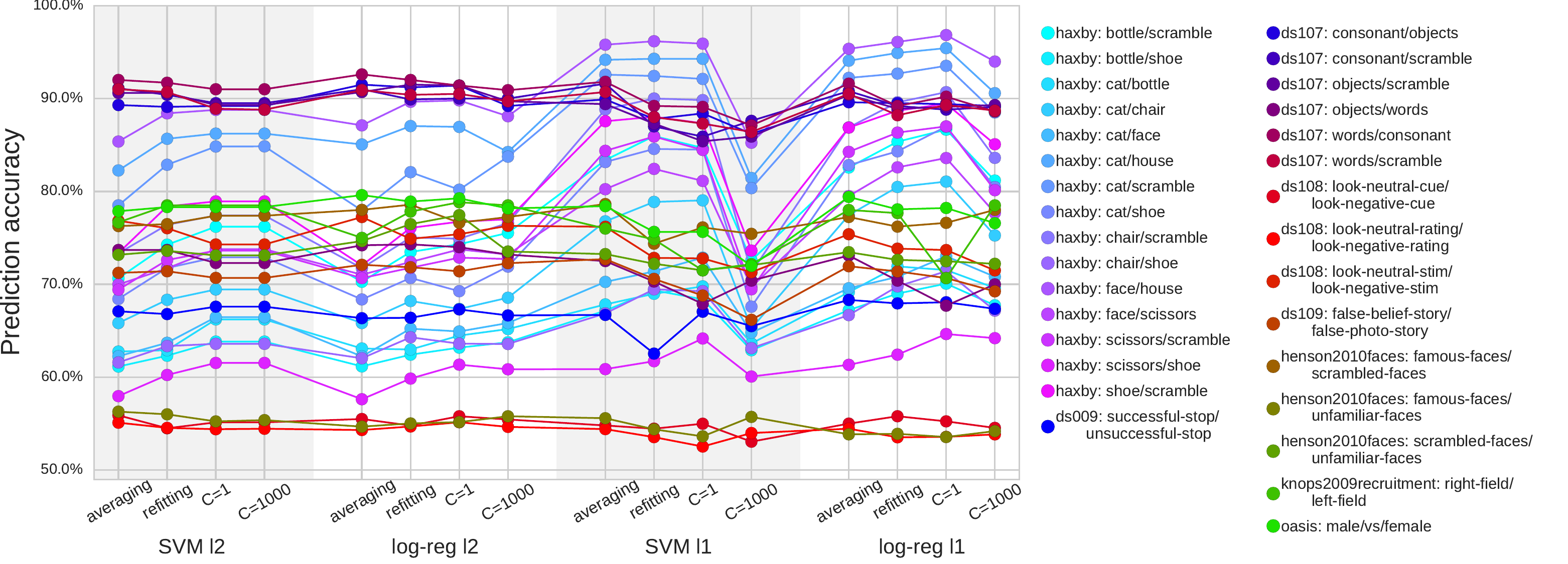}%
\caption{\textbf{Performance of each decoding strategy and each dataset}.
the plot is a "parallel coordinate plot": each lines denotes a dataset
and the different \emph{x} positions denote different decoding
strategies.
\label{fig:all_dataset}%
}
\end{minipage}
\end{figure}

The prediction accuracy results presented in the various figures are
differential effects removing the contribution of the dataset.
In figure \ref{fig:all_dataset}, we present for each decoding strategy
the prediction accuracy on all datasets.

We can see that the variations of prediction accuracy from one decoding
strategy to another are mostly reported across datasets: the various
lines are roughly parallel. One notable exception is SVM $\ell_1$ with
$C=1000$ for which some datasets show a strong decrease. Another, weaker,
variation is the fact that $\ell_1$ models tend to perform better on the
Haxby dataset (our source of intra-subject classification tasks). This
good performance of sparse models could be due to the intra-subject
settings: sparse maps are less robust to inter-subject variability.
However, the core messages of the paper relative to which
parameter-tuning strategy to use are applicable to intra and
inter-subject settings. For non-sparse models, using a large value of C
without parameter tuning is an overall safe choice, and for sparse
models, model averaging, refitting, and a choice of $C=1$ do not offer a
clear win, although model averaging is comparatively less variable.

\clearpage

\end{document}